\RequirePackage{fix-cm}
\documentclass[smallextended,natbib]{svjour3}       
\smartqed  
\usepackage{graphicx}
\usepackage{amsmath} 
\usepackage{amssymb} 
\usepackage{xcolor} 
\usepackage{booktabs} 
\usepackage{array} 
\usepackage{bbm} 
\usepackage{subcaption}
\usepackage[textsize=small]{todonotes}
\usepackage{multirow} 
\usepackage[boxed,linesnumbered]{algorithm2e}
\usepackage[hidelinks]{hyperref}
\definecolor{darkgreen}{rgb}{0,.8,0} 
\newcommand{\ChM}[2][]{\textcolor{black}{#2}}

\newcommand{\YG}[2][]{\textcolor{black}{#2}}

\journalname{Information Retrieval Journal}
\begin{document}

\title{Preference-based 
Interactive Multi-Document Summarisation
}
\subtitle{}


\author{Yang Gao        \and
        Christian M. Meyer \and 
        Iryna Gurevych 
}

\authorrunning{Y. Gao, C. M. Meyer, I. Gurevych} 

\institute{Yang Gao, Christian M. Meyer, Iryna Gurevych \at
              Ubiquitous Knowledge Processing Lab (UKP-TUDA) \\
                Department of Computer Science, Technische Universit\" at Darmstadt  \\
                \url{https://www.ukp.tu-darmstadt.de}
}

\date{Received: date / Accepted: date}

\maketitle

\begin{abstract}
Interactive NLP is a 
promising paradigm to close the gap between
automatic NLP systems and the human upper bound.
\emph{Preference-based interactive learning} has been 
successfully applied, but the existing methods
require several thousand interaction rounds even
in simulations with perfect user feedback.
In this paper, we study preference-based
interactive summarisation.
To reduce the number of interaction rounds, we
propose the \emph{Active Preference-based
ReInforcement Learning (APRIL)} framework.
APRIL uses \emph{Active Learning} to query the user, 
\emph{Preference Learning}
to learn a summary ranking function from 
the preferences, and 
neural \emph{Reinforcement Learning}
to efficiently search for the (near-)optimal summary.
\ChM[We compare APRIL with the state-of-the-art preference-based 
interactive technique in both simulation and real-user experiments;
r]{Our r}esults show that users can easily provide
reliable preferences over summaries\ChM[, 
and]{\ and that} APRIL outperforms the state-of-the-art \ChM[technique
in both setup]{preference-based interactive method in both simulation and real-user experiments}.

\end{abstract}

\section{Introduction}
\label{sec:introduction}

\emph{Interactive} Natural Language Processing (NLP) approaches 
that put the human in the loop 
gained increasing research interests recently
\citep{Amershi14,Gurevych:2017a,sr2018naacl}.
The user--system interaction enables personalised and 
user-adapted results by incrementally refining the underlying 
model based on a user's behaviour and by optimising the 
learning through actively querying for feedback and judgements.
Interactive methods can start with no or only few input 
data and adjust the output to the needs of human users.

\ChM[People have]{Previous research has} explored eliciting different 
forms of feedback from users in interactive NLP,
for example 
mouse clicks for information retrieval \ChM{\citep{Borisov18}},
\ChM{post-edits and} ratings for machine translation \ChM{\citep{Denkowski14,sr2018naacl}}, 
error markings for semantic parsing \citep{DBLP:journals/corr/abs-1811-12239},
bigrams for summarisation
\citep{avinesh_meyer2017_interactive_doc_sum},
and preferences
for translation \citep{sr2018acl}.
%
\YG{\ChM[Research has suggested that, in controlled experiments,]{Controlled experiments suggest that }asking for preferences places a lower cognitive burden
on the human subjects than asking for \ChM{absolute}
ratings or categorised labels
\citep{thurstone27comparitive,kendall1948rank,kinsley2010perference}. 
But it remains unclear whether people can easily provide
reliable preferences over summaries.}
In addition, preference-based interactive NLP faces the \emph{high sample complexity
problem}: a preference is a binary decision and hence
only contains a single bit of information,
so the NLP systems usually need to
elicit a large number of preferences from the 
users to improve their performance.
\ChM[\cite{sokolov_etal16interactive_nlp}
have shown that their translation system]{For example, the machine translation system by \citet{sokolov_etal16interactive_nlp}}
needs to collect hundreds of thousands
of preferences from a simulated user
before it converges.

Collecting such large amounts of user inputs and \ChM[use]{using} them
to train a ``one-fits-all'' model might be 
feasible for tasks such as machine translation, because the learnt model 
can generalise to many unseen texts.
However, for highly subjective tasks, such as document summarisation, 
this procedure is not effective, since the notion of importance 
is specific to a certain topic or user.
For example, the information that Lee Harvey Oswald shot president 
Kennedy might be important when summarising the assassination, 
but less important for a summary on Kennedy's childhood.
\ChM{Likewise, a user who is not familiar with the assassination might consider the information more important than a user who is analysing the political backgrounds for many years.}
Therefore, we aim at an interactive system that adapts a model 
for a given topic and user context based on user feedback -- 
instead of training a single model across all users and topics, 
which hardly fits anyone's needs perfectly.
In this scenario, it is essential to overcome the high 
sample complexity problem and learn to adapt the model using 
a minimum of user interaction.

\begin{figure}
  \centering
  \begin{subfigure}[b]{0.6\textwidth}
    \includegraphics[width=\textwidth]{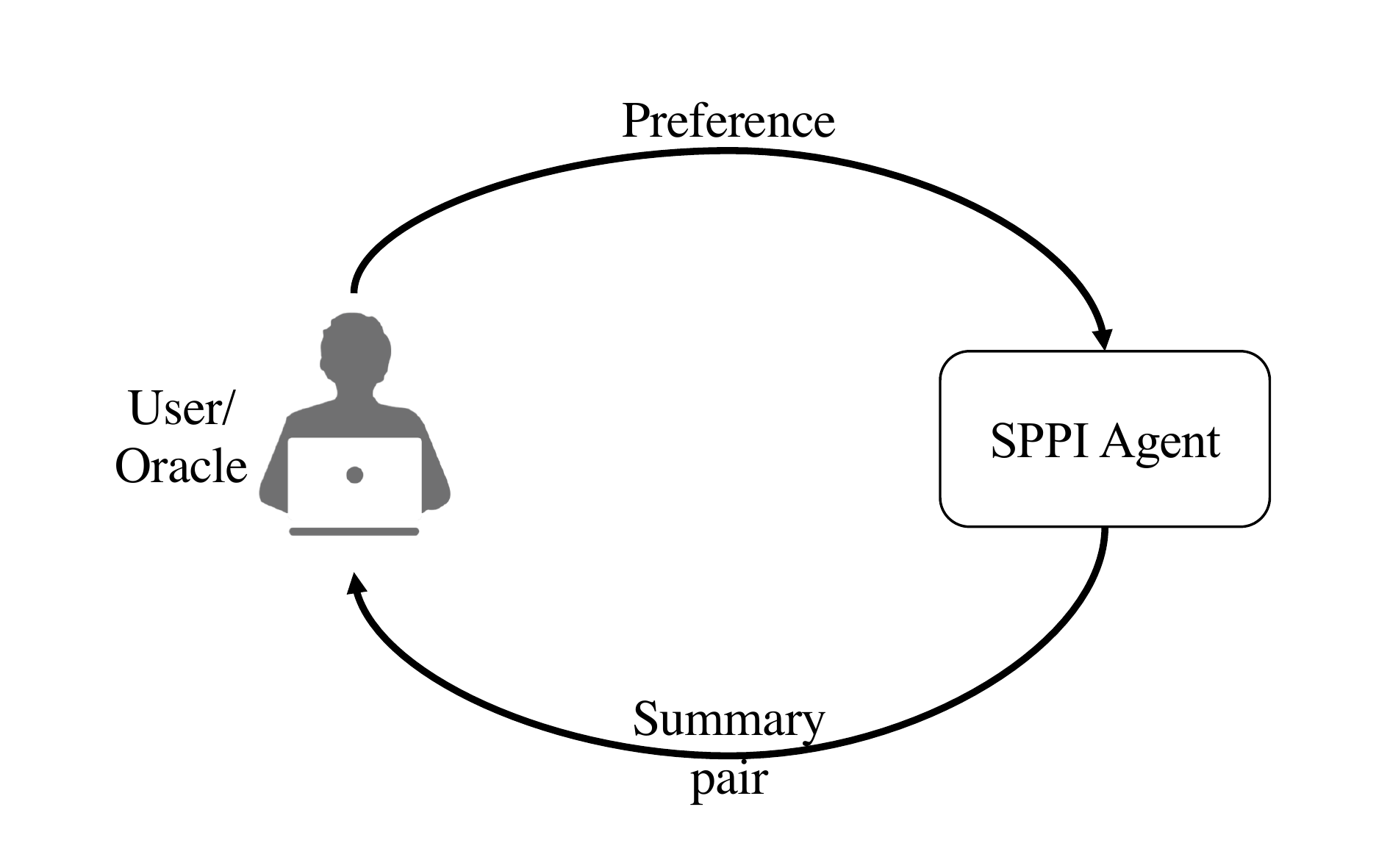}
    \caption{SPPI workflow}
    \label{subfig:workflow:sppi}
  \end{subfigure}
  \begin{subfigure}[b]{0.89\textwidth}
    \includegraphics[width=\textwidth]{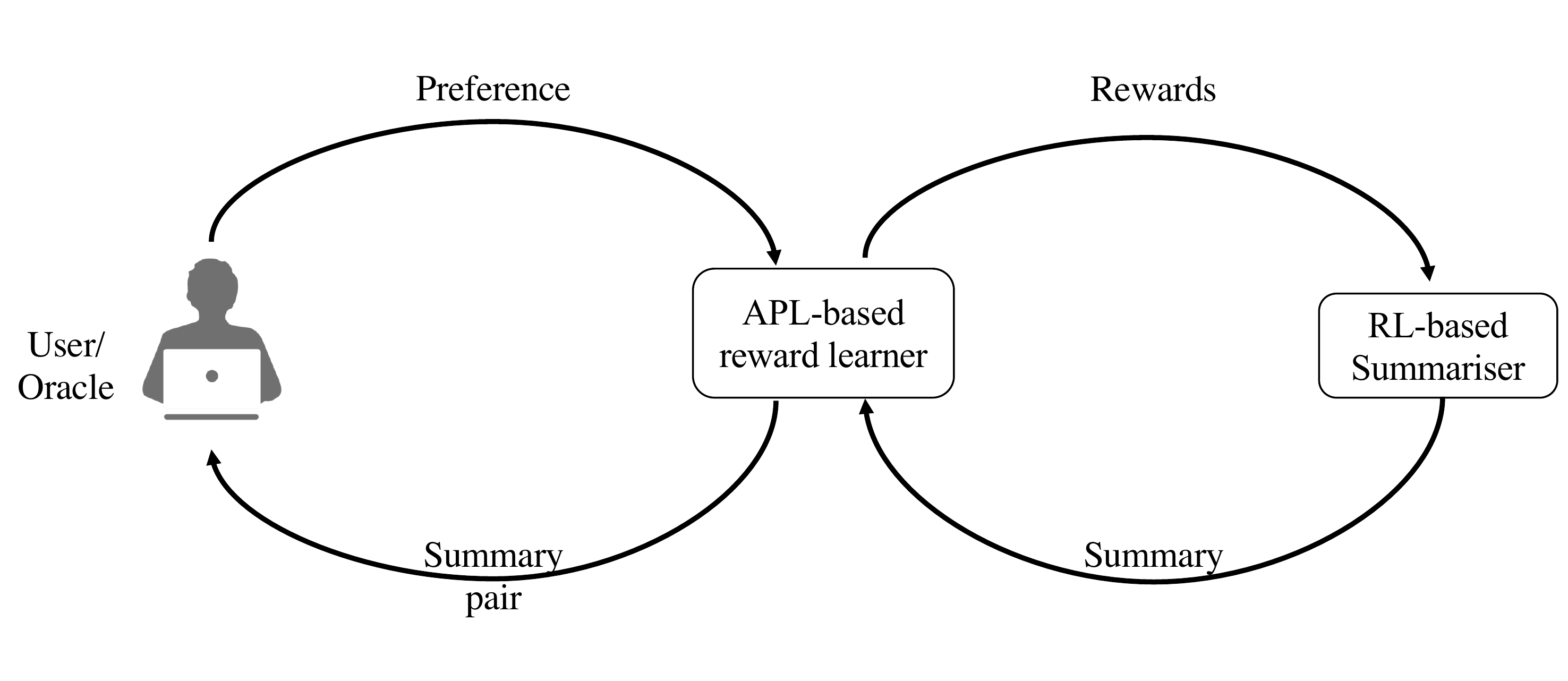}
    \caption{APRIL workflow}
    \label{subfig:workflow:april}
  \end{subfigure}
  \caption{\YG[Workflows of SPPI (a) and APRIL (b).]{SPPI (a)
  directly uses the collected preferences to ``teach'' its summary-generator,
  while APRIL (b) learns a reward function as the proxy of the user/oracle, and uses the
  learnt reward to ``teach'' the RL-based summariser.
  } }
  \label{fig:workflows}
\end{figure}

In this \ChM[paper]{article}, we propose the \emph{Active Preference-based 
ReInforcement Learning (APRIL)} framework\footnote{\YG{
We first introduced APRIL in \citep{emnlp18}. 
Towards the end of \S\ref{sec:introduction} we discuss how this article
substantially extends our previous work.}}.
Our core research idea is to split the 
\YG[machine learning workflow]{preference-based interactive
learning process} into two stages.
First, we estimate the user's ranking over candidate summaries 
using \emph{active preference learning} (APL) in an interaction loop.
Second, we use the learnt ranking to guide a neural 
\emph{reinforcement learning} (RL) agent to search for the (near-)optimal summary. 
The use of APL allows us to maximise the information gain from
a small number of preferences, helping to 
reduce the sample complexity.
Fig.~\ref{fig:workflows} shows this general idea in comparison 
to the \YG{state-of-the-art preference-based interactive NLP paradigm,
\emph{Structured Prediction from Partial Information (SPPI)}
\citep{DBLP:conf/nips/SokolovKRL16,DBLP:conf/acl/KreutzerSR17}.
}
\ChM{In \S\ref{sec:backgrond}, we discuss the technical background of RL, 
preference learning and SPPI, before we introduce our 
solution APRIL in \S\ref{sec:pb_doc_sum}.}

\ChM{We apply APRIL to the
\emph{Extractive Multi-Document Summarisation (EMDS)} task.
Given a cluster of documents on the same topic,
an EMDS system needs to extract important sentences
from the input documents to generate 
a summary complying with a given length requirement 
that fits the needs of the user and her/his task.}
For the first time,
we provide evidence for the efficacy of preference-based interaction in 
EMDS based on \ChM[two user studies]{a user study}, in which we 
measure the usability and the noise of preference feedback, 
yielding a mathematical model we can use for simulation and 
for analysing our results\ChM{\ (\S\ref{sec:pref_summary})}.
To evaluate \ChM[the proposed model]{APRIL}, we then perform experiments on standard
EMDS benchmark datasets. 
We compare the effectiveness of multiple APL and RL algorithms
and select the best algorithms \ChM[to implement APRIL]{for our full system}.
\ChM[Then, w]{W}e compare APRIL to SPPI\ChM{\ and non-interactive methods},
in both simulation\ChM{\ (\S\ref{sec:experiments})} and real-user experiments\ChM{\ (\S\ref{sec:final_user_study})}.
\ChM[R]{Our r}esults suggest that with only \ChM[10]{ten} rounds of \ChM{user} interaction,
APRIL produces summaries better
than those produced by both non-interactive 
\ChM[baselines]{methods} and SPPI.

This work extends our earlier work 
\citep{emnlp18} in three aspects.
(i) We \ChM[added the]{present a new user} study on the reliability and \ChM[user-friendliness]{usability}
of the preference-based interaction 
(\S\ref{sec:pref_summary}). Based on this study,  \ChM{we propose}
a realistic simulated user\ChM[ is proposed]{}, which
is used in our \ChM[simulation experiments]{experiments}.
(ii) \ChM[M]{We evaluate m}ultiple new \ChM[PAL]{APL} strategies and a novel 
neural RL algorithm, \ChM[ were proposed]{} and compare\ChM[d]{\ them}
with the counterpart methods used in \ChM[\citep{emnlp18}
(\S\ref{subsec:stage2})]{ \citet{emnlp18}}.
\YG[The performance
of APRIL is further advanced with the help of these
new APL and RL algorithms]{The use of these new algorithms
further boost the efficiency and performance of APRIL}\ChM{\ (\S\ref{sec:experiments})}.
(iii) \ChM[A]{We conduct a}dditional user studies \ChM[were performed ]{}to compare
APRIL with both non-interactive baselines and SPPI
under more realistic settings (\S\ref{sec:final_user_study}).
APRIL can be applied to a wide range of other NLP tasks\ChM{, including machine translation, semantic parsing and information exploration}.
All source code and experimental setups
can be found in \url{https://github.com/UKPLab/irj-neural-april}. 

\section{Related Work}
\label{sec:related_work}

\paragraph{SPPI.}
The method most similar to ours is SPPI \citep{DBLP:conf/nips/SokolovKRL16,DBLP:conf/acl/KreutzerSR17}.
The core of SPPI is a policy-gradient RL 
algorithm, which receives rewards derived from the preference-based
feedback. It maintains a policy that approximates the utility
of each candidate output and selects the higher-utility candidates
with higher probability. As discussed in \S\ref{sec:introduction},
SPPI suffers heavily from the high sample complexity problem.
We will present the technical details of SPPI \ChM{in \S\ref{subsec:sppi} }and compare it
to APRIL in \ChM[ \S\ref{subsec:sppi}]{\S\ref{sec:experiments} and \S\ref{sec:final_user_study}}.

\paragraph{Preferences.}
The use of preference-based feedback in NLP attracts increasing
research interest. \citet{conf/naacl/zopf18pref} 
learns a sentence ranker from human preferences on 
sentence pairs, which can be used to evaluate
the quality of summaries, by counting how many high-ranked sentences
are included in a summary. 
\citet{edwin2018tacl} develop an improved \emph{Gaussian process preference learning} 
\citep{DBLP:conf/icml/ChuG05} algorithm to learn an argument convincingness 
ranker from noisy preferences.
Unlike these methods that focus on learning a ranker from
preferences, we focus on using preferences to generate
better summaries.
\citet{sr2018acl} ask real users to provide cardinal
(5-point ratings) and ordinal (pairwise preferences) feedback
over translations, and use the collected data
to train an off-policy RL to improve the translation quality. 
Their study suggests that the inter-rater agreement 
for the cardinal and ordinal
feedback is similar. However, they do not measure or 
consider the influence of the
questions' difficulties on the agreement, which we find significant for EMDS 
(see \S\ref{sec:pref_summary}).
In addition, their system is not interactive, but \ChM[instead
uses log data, unlike our system that actively
queries users to reduce the sample complexity]{uses log data instead of actively querying users}. 

\paragraph{Interactive Summarisation.}
The iNeATS \citep{leuski2003ineats}
and IDS
\citep{joneetal2002interactive_doc_sum} systems
allow users to tune several parameters 
(e.g., size, redundancy, focus) 
to customise the produced summaries.
Further work presents automatically derived summary templates 
\citep{orasan:2003,orasan:2006} or hierarchically ordered summaries
\citep{DBLP:conf/acl/ChristensenSBM14,DBLP:conf/emnlp/ShapiraRAABD17}
allowing users to drill-down from a general overview to detailed information.
However, these systems do not employ the users' 
feedback to update their internal 
summarisation models. 
\cite{avinesh_meyer2017_interactive_doc_sum} 
propose an interactive EMDS system
\ChM[requesting]{that asks users} to label important bigrams
within candidate summaries.
Given the important bigrams, they use \emph{integer linear
programming} to optimise important bigram
coverage in the summary.
In simulation experiments, their system can achieve 
near-optimal performance in ten rounds of interaction, 
collecting up to 350 important bigrams. 
However, labelling important bigrams 
is a large burden on the users, 
as users have to read through many 
potentially unimportant bigrams\ChM{\ (see \S\ref{sec:pref_summary})}. 
Also, they assume that the users' feedback is always perfect.

\paragraph{Reinforcement Learning.}
RL has been applied 
to both extractive and abstractive summarisation in recent years
\citep{ryang2012emnlp,rioux2014emnlp,DBLP:conf/acl/GkatziaHL14,ig2015rl,paulus_etal2017deep_rl_abs_summary,DBLP:conf/naacl/PasunuruB18,DBLP:conf/emnlp/KryscinskiPXS18}. 
Most existing RL-based document summarisation systems
either use heuristic functions (e.g., 
\citealp{ryang2012emnlp,rioux2014emnlp}), which 
do not rely on reference summaries, or 
ROUGE scores requiring reference summaries as the rewards for RL 
\citep{paulus_etal2017deep_rl_abs_summary,DBLP:conf/naacl/PasunuruB18,DBLP:conf/emnlp/KryscinskiPXS18}.
However, neither ROUGE nor the heuristics-based
rewards can precisely reflect real users'
requirements on summaries \citep{chaganty2018price};
hence, using these imprecise rewards can
severely mislead the RL-based summariser. 
The quality of the rewards has been
recognised as the bottleneck for RL-based
summarisation systems \citep{DBLP:conf/emnlp/KryscinskiPXS18}.
Our work \ChM[tries to learn the good rewards]{learns how to give good rewards}
from users' preferences.
In this work, we assume that our system
has no access to the reference summaries, 
but can query a user for preferences\ChM{\ over summary pairs}.

Some RL work directly uses the users' ratings as rewards.
\citet{DBLP:conf/emnlp/NguyenDB17} employ user
ratings on translations as rewards when training an RL-based
encoder-decoder translator. 
However, eliciting ratings on summaries is very expensive
as users have high variance in their ratings
of the same summary \citep{chaganty2018price}\ChM{, which is why we consider preference-based feedback and a learnt reward surrogate}.

\paragraph{Preference-based RL (PbRL)}
is a recently proposed paradigm
at the intersection of preference learning, RL,
active learning (AL) and inverse RL 
\citep{DBLP:journals/jmlr/WirthANF17}.
Unlike \emph{apprenticeship learning} \citep{DBLP:conf/acl/DethlefsC11}
which requires the user to demonstrate (near-)optimal 
sequences of actions (called \emph{action trajectories}),
PbRL only asks for the user's preferences (either
partial or total order) on several action trajectories.
\citet{wirth_etal2016aaai_pbrl} apply PbRL 
to several simulated robotics tasks.
They show \ChM{that} their method
can achieve near-optimal performance 
by interacting with a simulated perfect user for 15--40 rounds. 
\citet{christiano2017pbrl} use PbRL in training 
simulated robotics tasks, Atari-playing agents and a
simulated back-flipping agent
by collecting feedback from both simulated oracles and real 
crowdsourcing workers.
They find that human feedback
can be noisy and partial 
(i.e., capturing only a fraction of the true reward),
but that it is much easier for people to provide consistent 
comparisons than consistent absolute scores in their robotics use case.
\ChM{In \S\ref{sec:pref_summary}, we evaluate this for document summarisation.}

However, \ChM[their approach]{the approach by \citet{christiano2017pbrl}} fails to obtain satisfactory 
results in some \ChM{robotics} tasks even after 5,000 interaction rounds.
In a follow-up work, \citet{DBLP:conf/nips/IbarzLPILA18}
elicit demonstrations from experts,
use the demonstrations to pre-train a model
with imitation learning techniques, and 
\ChM{successfully} fine-tune the pre-trained model with PbRL.
In \ChM[document summarisation]{EMDS}, \ChM[reference extractive summaries]{extractive reference summaries}
might be viewed as demonstrations, but they are expensive to collect and 
not available in popular summarisation
corpora (e.g., the DUC datasets).
\ChM{APRIL does not require demonstrations, but learns a reward function based on user preferences on entire summaries, which is then used to train an RL policy.}


\section{Background}
\label{sec:backgrond}
In this section, we recap necessary details of RL (\S\ref{subsec:rl}),
preference learning (\S\ref{subsec:pref_learn}) and SPPI
(\S\ref{subsec:sppi}). We adapt them to the EMDS use case,
so as to lay the foundation for APRIL. 
To ease the reading, we summarise the notation used in \ChM[this section at]{the remaining article in}
Table~\ref{table:notations}.

\begin{table}
    \centering
    \begin{tabular}{l p{.7\linewidth}}
    \toprule
    Notation & Description \\ 
    \midrule
    $x \in \mathcal{X}$ & a document cluster $x$ from the set of all possible inputs $\mathcal{X}$ \\
    $y \in \mathcal{Y}_x \subseteq S$ & a summary $y$ from the set of all legal summaries $\mathcal{Y}_x$ for $x \in \mathcal{X}$ \\
    $M_x = (S,A,P,R,T)$ & 
    MDP of the EMDS task for $x \in \mathcal{X}$: states $S$,
    actions $A$, transition function $P$, reward function $R$ and terminal states $T \subseteq S$ \\
    $R(y)$ & 
    the reward of summary $y$ in $M_x$ \\
    $\pi(y)$ 
    & policy in RL: the probability of selecting summary $y$ \ChM[for $x$]{in $M_x$} \\
    $\pi(\langle y_i, y_j \rangle; w)$ &  policy in SPPI, parameterised by $w$:
    the probability of presenting pair $\langle y_i, y_j \rangle$ to the oracle
    (Eq.~\eqref{eq:sr_query})\\
    $U^*_x$ & the ground-truth utility function on $\mathcal{Y}_x$ \\
    %
    $\hat{U}_x$ & the approximation of $U^*_x$ \\
    $\Delta_{U_x}(y_i,y_j)$ &  $U_x(y_i) - U_x(y_j)$,
    where $U_x$ is a utility function on $\mathcal{Y}_x$ \\
    $\sigma^*_x$ &
    the ranking function on $\mathcal{Y}_x$\ChM[; $\sigma^*_x(y)$ returns
    the number of summaries in $\mathcal{Y}_x$ that has lower
    $U^*_x$ value than $y$]{\ induced by $U_x^*$} (Eq.~\eqref{eq:sigma}) \\
    $\hat{\sigma}_x$ & the approximation of  $\sigma^*_x$ \ChM{induced by $\hat{U}_x$} \\
    $p_x(\langle y_i, y_j \rangle)$ & the preference direction function,
    which returns 1 if the oracle/user prefers $y_i$ over $y_j$ for $x$ \\
    $\mathcal{R}_x^\mathrm{RL}$ & the objective function in RL (Eq.~\eqref{eq:rl_obj})\\
    $\mathcal{R}_x^\mathrm{BT}$ & the objective function in preference learning (Eq.~\eqref{eq:mal_bt})\\
    $\mathcal{R}_x^\mathrm{SPPI}$ & the objective function in SPPI (Eq.~\eqref{eq:sr_loss})\\
    \bottomrule
    \end{tabular}
    \caption{Overview of the notation used in this article
    }
    \label{table:notations}
\end{table}

\subsection{Reinforcement Learning}
\label{subsec:rl}
RL amounts to algorithms for efficiently searching optimal solutions
in \emph{Markov Decision Processes (MDPs)}. MDPs are widely used
to formulate \emph{sequential decision-making problems}\ChM[,
which EMDS falls into: in EMDS, the summariser has to sequentially select
sentences from original documents to add to the draft summary]{}.
%
\ChM{Let $\mathcal{X}$ be the input space and let $\mathcal{Y}_x$ be the set of all possible outputs for input $x \in \mathcal{X}$.}
An episodic MDP is a tuple $\ChM{M_x} = (S,A,P,R,T)$\ChM[.]{\ for input $x \in \mathcal{X}$, where} $S$ is the set of 
\emph{states}, $A$ is the set of \emph{actions} and 
$P\colon S \times A \to S$ 
is the \emph{transition function} with $P(s,a)$ giving
the next state after performing action $a$ in state $s$.
$R\colon S \times A \to \mathbb{R}$
is the \emph{reward function} with $R(s,a)$ giving the immediate reward for
performing action $a$ in state $s$. 
$T \subseteq S$ is the set of 
\emph{terminal states}; visiting a terminal state terminates the
current episode. 

EMDS can be formulated  as episodic MDP, as the summariser has to sequentially select
sentences from the original documents to add to the draft summary. 
Our MDP formulation of EMDS matches previous approaches 
by \citet{ryang2012emnlp} and \citet{rioux2014emnlp}:
%
 $x \in \mathcal{X}$ is a cluster of documents and
$\mathcal{Y}_x$ is the set of all legal summaries for cluster $x$ (i.e., all permutations of sentences in $x$ that fulfil the given summary length constraint).
%
\ChM{In the MDP $M_x$ for document cluster $x \in \mathcal{X}$,}
$S$ includes all possible draft summaries of any length (i.e., $\mathcal{Y}_x \subseteq S$). The action set
$A$ includes two types of actions: \emph{concatenate} a 
sentence in $x$ to the current draft summary,
or \emph{terminate} the draft summary construction. The transition function $P$ is trivial in EMDS, because 
given the current draft summary
and an action, the next state can be easily \ChM[inferred]{identified} as the draft summary plus the selected sentence or as a terminating state.
The reward function $R$ returns an evaluation score of the summary
once the action \emph{terminate} is performed; otherwise it returns 0
because the summary is still under construction and thus not ready
to be evaluated (so-called \emph{delayed rewards}). 
Providing non-zero rewards before the action \emph{terminate}
can lead to even worse result, as reported by \citet{rioux2014emnlp}.
The terminal states set $T$ includes all states corresponding to summaries exceeding the given length requirement and an \emph{absorbing state} $s_T$.
By performing action \emph{terminate}, the agent will be transited to
$s_T$ regardless of its current state, i.e. $P(s,a) = s_T$ for all $s \in S$ 
if $a$ is \emph{terminate}.

A \emph{policy} $\pi\colon S \times A \to \mathbb{R}$ in an MDP \ChM{$M_x$}
defines how actions are selected: $\pi(s,a)$ is the probability
of selecting action $a$ in state $s$.
\ChM{Note that in many sequential decision-making tasks, $\pi$ is learnt across all inputs $x \in \mathcal{X}$. However, for our EMDS use case, we learn an \emph{input-specific} policy for a given $x \in \mathcal{X}$ in order to reflect the subjectivity of the summarisation task introduced in \S\ref{sec:introduction}.}
%
We let $\mathcal{Y}_\pi$
be the set of all possible summaries a policy $\pi$ can construct
in document cluster $x$.  $\pi(y)$ denotes the probability of policy $\pi$
for generating a summary $y$ in $x$.
Likewise, $R(y)$ denotes the accumulated
reward received by building summary $y$.
Finally, the expected reward of performing $\pi$ is:
\begin{align}
\mathcal{R}_x^\mathrm{RL}(\pi) & = \mathbb{E}_{y \in 
\mathcal{Y}_{\pi}} R(y)
 = \sum\limits_{y \in \mathcal{Y}_{\pi}} \pi(y)R(y).
\label{eq:rl_obj}
\end{align}
The goal of an MDP is to find the optimal policy $\pi^*$
that has the highest expected reward: $\pi^* =\arg\max_{\pi}\mathcal{R}^\mathrm{RL}(\pi)$.

\subsection{Preference Learning}
\label{subsec:pref_learn}
For a document cluster $x \in \mathcal{X}$ and its
legal summaries set $\mathcal{Y}_x$,
we let $U^*_x:\mathcal{Y}_x \to \mathbb{R}$ 
be the ground-truth \emph{utility function} measuring the quality
of summaries in $\mathcal{Y}_x$.
We additionally assume that no two items 
in $\mathcal{Y}_x$ have the same $U^*_x$ value.
Let $\sigma^*_x$ be the ascending ranking induced by $U^*_x$:
for $y \in \mathcal{Y}_x$,
\begin{align}
\sigma^*_x(y) = \sum_{y' \in \mathcal{Y}_x} \mathbbm{1}[U^*_x(y') < U^*_x(y)], 
\label{eq:sigma}
\end{align}
where $\mathbbm{1}$ is the indicator function.
In other words, $\sigma^*_x(y)$ gives the rank of $y$
among all elements in $\mathcal{Y}_x$ with respect to $U^*_x$.
The goal of preference learning is to 
\ChM[reconstruct or approximate]{approximate} $\sigma^*_x$
from the pairwise preference\ChM{s} on some elements in $\mathcal{Y}_x$.
The preferences are provided by an \emph{oracle}.

The Bradley-Terry (BT) model \citep{bradley1952rank}
is a widely used preference learning model, which
approximates the ranking $\sigma^*_x$ by approximating
the utility function $U^*_x$:
Suppose we have observed $N$ preferences: 
$\{p_x(\langle y_{1,1},y_{1,2} \rangle), \cdots, 
p_x( \langle y_{N,1},y_{N,2} \rangle)\}$, 
where $y_{i,1}, y_{i,2} \in \mathcal{Y}_x$ are the
summaries presented to the oracle in the $i^\mathrm{th}$ round,
and $p_x$ indicates the \emph{preference direction} of 
the oracle: 
$p_x = 1$ if the oracle prefers $y_{i,1}$ over $y_{i,2}$,
and $p_x = 0$ otherwise.
The objective in BT is to maximise the following likelihood function:
\begin{align}
\mathcal{R}^\mathrm{BT}_x(w)=  \sum_{i \in N} 
     \quad[\;&  p_x(\langle y_{i,1},y_{i,2} \rangle) \,\log \mathcal{P}_x(y_{i,1}, y_{i,2};w) 
     \nonumber \\
     &  +  p_x(\langle y_{i,2},y_{i,1} \rangle) \,\log \mathcal{P}_x(y_{i,2}, y_{i,1};w) \, ],
      \label{eq:mal_bt}
\end{align}
where 
\begin{align}
\mathcal{P}_x(y_i, y_j;w) = 
\frac{1}{1+\exp[\hat{U}_x(y_j;w)-\hat{U}_x(y_i;w)]};
\label{eq:bt}
\end{align}
$\hat{U}_x$ is the approximation of $U^*_x$ parameterised by $w$,
which can be learnt by any function approximation techniques,
e.g. neural networks or linear models.
By maximising Eq.~\eqref{eq:mal_bt}, the resulting $w$ will be used
to obtain $\hat{U}_x$, which in turn can be used to induce
the approximated ranking function 
$\hat{\sigma}_x \colon \mathcal{Y}_x \rightarrow \mathbb{R}$.

\subsection{The SPPI Framework}
\label{subsec:sppi}
SPPI can be viewed as a combination of RL and preference learning.
%
For an \ChM[input document cluster]{input} $x \in \mathcal{X}$,
the objective of SPPI is to maximise 
\begin{align}
 \mathcal{R}^\mathrm{SPPI}_x(w) & = \mathbb{E}_{\pi(\langle y_{i},y_{j} \rangle; w)}[
p_x(\langle y_{i},y_{j} \rangle)] \nonumber \\ 
& =  
\sum_{y_i,y_j \in \mathcal{Y}_x}  \;
\pi(\langle y_t,y_k \rangle; w) \cdot
p_x(\langle y_i,y_j \rangle)
\label{eq:sr_loss},
\end{align}
where $p_x$ is the same preference direction function
as in preference learning (\S\ref{subsec:pref_learn}).
$\pi$ is a policy that decides the probability of presenting
a pair of summaries to the oracle:
\begin{align}
\pi(\langle y_{i},y_{j} \rangle;w) 
& = \frac{\exp[\hat{U}_x(y_{i};w)-\hat{U}_x(y_{j};w)]}
{\sum\limits_{y_p,y_q \in \mathcal{Y}_x}
\exp[\hat{U}_x(y_p;w) - \hat{U}_x(y_q;w)]} 
\label{eq:sr_query}.
\end{align}
In line with preference learning, $\hat{U}_x$ is the utility function
for estimating the quality of summaries, parameterised by $w$.  
The policy $\pi$ samples the pairs with larger utility gaps 
with higher \ChM[probabilities]{probability}; as such, both ``good''
and ``bad'' summaries have \ChM{the }chance to be presented
to the oracle\ChM[, which]{\ and thus} encourages the exploration of the \ChM[summaries]{summary space}.
To maximise Eq.~\eqref{eq:sr_loss}, SPPI uses gradient ascent
to update $w$ incrementally. Algorithm \ref{alg:sr} presents
the pseudo code of our adaptation of SPPI to EMDS.

\begin{algorithm}[t]
\SetKwInOut{Input}{Input}
\SetKwInOut{Output}{Output}
  \Input{sequence of learning rates $\gamma$; query budget $N$; 
  document cluster $x$}
	initialise $w_0$\;
  \While{$i=0, \ldots, N-1$}{
  	sampling $ y_{i,1},y_{i,2} $ using $\pi(\cdot;{w_i})$ (Eq.~\ref{eq:sr_query})\;
    get preference $p_x(\langle y_{i,1},y_{i,2} \rangle)$\;
    $w_{i+1} := w_i + \gamma \nabla_w \mathcal{R}^\mathrm{SPPI}_x(w_i)$ (Eq.~\ref{eq:sr_loss})\;
    }
  \Output{$y = \arg \max_y w_N \cdot \phi(y|x)$}
  \caption{Adaptation of SPPI \citep[ Alg.~1]{DBLP:conf/acl/KreutzerSR17} for preference-based EMDS.
  \label{alg:sr}}
\end{algorithm}

Note that the objective function in SPPI (Eq. \eqref{eq:sr_loss})
and the expected reward function in RL (Eq. \eqref{eq:rl_obj})
\ChM[are in similar forms]{have a similar form}: 
if we view the preference direction function
$p_x$ in Eq.~\eqref{eq:sr_loss} as a reward function, 
we can \ChM[view]{consider} SPPI as an RL problem.
The major difference between SPPI and RL is that 
the policy in SPPI selects pairs (Eq.~\eqref{eq:sr_query}),
while the policy in RL selects single summaries
(see \S\ref{subsec:rl}).
\ChM[W]{For APRIL, w}e will exploit \ChM[their]{this} connection
to propose our new objective function and learning paradigm.

\section{The APRIL Framework}
\label{sec:pb_doc_sum}
SPPI suffers from the high sample complexity problem, 
which we \ChM[believe it is largely due to two]{attribute to two major} reasons:
First, the policy $\pi$ in SPPI (Eq.~\eqref{eq:sr_query}) 
is good at \ChM[identifying]{distinguishing}
the ``good'' summaries from the ``bad'' ones, but poor
at selecting the ``best'' summaries from ``good'' summaries,
because it only queries the summaries with large quality gaps.
Second\ChM[ is its]{, SPPI makes} inefficient use of the collected preferences:
\ChM[Note that a]{A}fter each round of interaction, 
SPPI performs one step of the policy gradient update\ChM[.
SPPI]{, but} does not generalise or re-use the collected preferences\ChM[,]{. This}
potentially \ChM[wasting the]{wastes} expensive \ChM{user} information.
To alleviate these two problems, we exploit the connection
between SPPI, RL and preference learning and propose the
APRIL framework detailed in this section.

Recall that in EMDS, the goal is to find the optimal summary for a given
document cluster $x$, namely the summary that is preferred over all
other possible summaries in $\mathcal{Y}_x$
according to $\sigma^*_x$. 
Based on this understanding\ChM[,]{\ and}
in line with the RL formulation of EMDS \ChM[described in Section ]{from \S}\ref{subsec:rl}, 
we define a new expected reward function $\mathcal{R}_x^\mathrm{APRIL}$ 
for policy $\pi$ as follows:
\begin{align}
 \mathcal{R}^\mathrm{APRIL}_x(\pi) & = \mathbb{E}_{y_j \sim \pi} \bigg[
\sum\limits_{y_i\in \mathcal{Y}_x} p_x(\langle y_i,y_j \rangle) \bigg] \nonumber \allowdisplaybreaks
\\
& = 
\sum\limits_{y_j\in \mathcal{Y}_x(\pi)} \pi(y_j)
\sum\limits_{y_i\in \mathcal{Y}_x} p_x(\langle y_i,y_j \rangle) 
\nonumber \allowdisplaybreaks
\\
& = 
\sum\limits_{y\in \mathcal{Y}_x(\pi)} \pi(y) \; \sigma_x^*(y),
\label{eq:our_obj}
\end{align}
Note that $p_x(\langle y_i,y_j \rangle )$ equals 1 if 
$y_j$ is preferred over $y_i$ and equals 0 otherwise
(see \S\ref{subsec:pref_learn}). Thus, 
$\sum_{y_i\in \mathcal{Y}_x} p_x(\langle y_i,y \rangle)$
counts the number of summaries that are less-preferred
than \ChM{summary }$y$, and hence equals $\sigma^*_x(y)$ 
(see Eq.~\ref{eq:sigma}).
Policy that can maximise this new objective function
will select summaries with highest rankings, hence
outputs the optimal summary.

\begin{figure}
    \centering
    \includegraphics[width=0.85\textwidth]{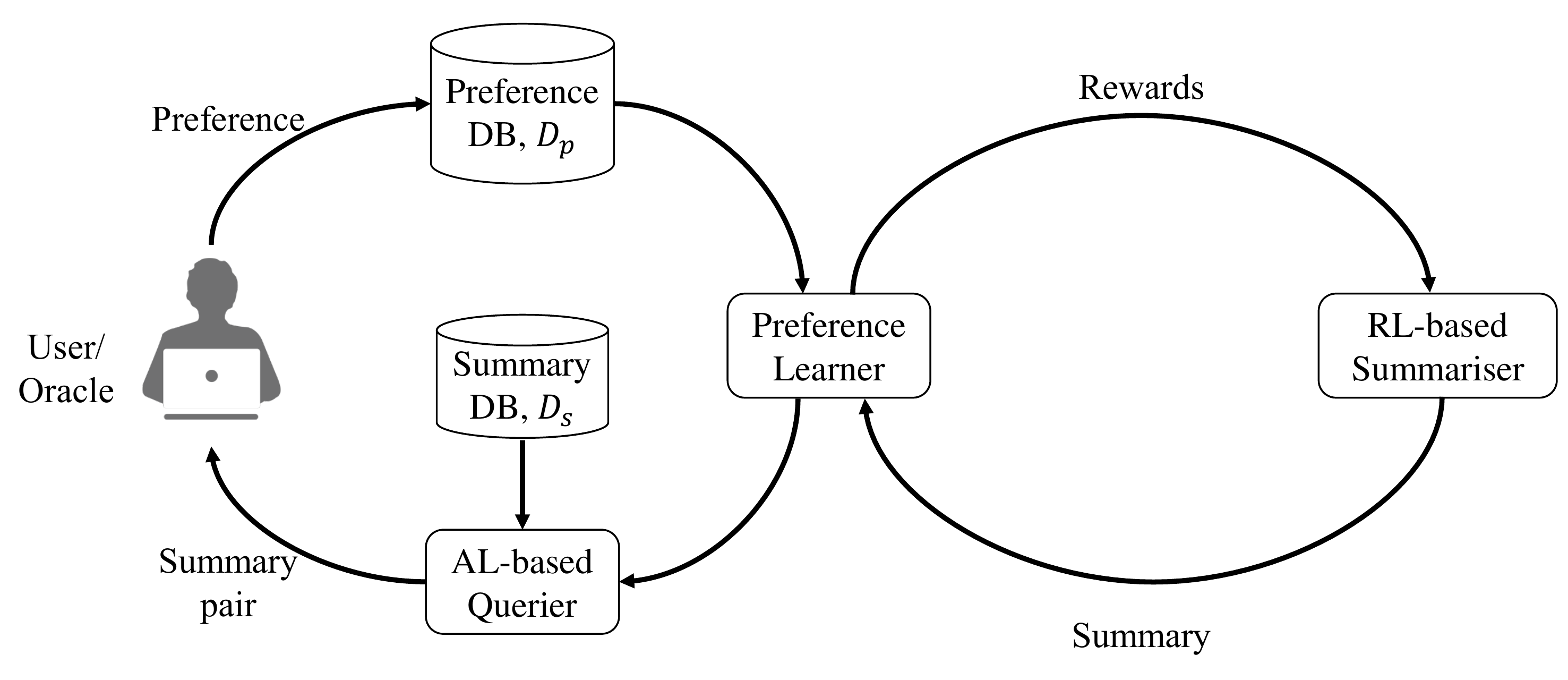}
    \caption{Detailed workflow of APRIL (extended version
    of the workflow presented in Fig. \ref{subfig:workflow:april})}
    \label{fig:april_workflow}
\end{figure}

This new objective function decomposes the learning problem
into two \ChM[sub-tasks]{stages}: 
(i) approximating the ranking function $\sigma^*_x$, and 
(ii) based on the approximated ranking function, searching for
the optimal policy that can maximise the new objective function.
These two \ChM[sub-tasks]{stages} can be solved by (active) preference learning and
reinforcement learning, respectively, and they constitute
our APRIL framework, illustrated in Figure~\ref{fig:april_workflow}. 

\subsection{\ChM[Stage1]{Stage 1}: Active Preference Learning}
\label{subsec:stage1}
For an input document cluster $x \in \mathcal{X}$,
the task in the first stage of APRIL is to obtain $\Hat{\sigma}_x$,
the approximated ranking function on $\mathcal{Y}_x$
by collecting a small number of preferences from the oracle.
It involves four major components:
a \emph{summary Database (DB)} 
storing the summary candidates,
an \emph{AL-based Querier}
that selects candidates from the Summary DB to present
to the user, 
a \emph{Preference DB} storing the preferences collected
from the user, and a \emph{Preference Learner} that
learns $\Hat{\sigma}_x$ from the preferences.
The left cycle in Fig.\ \ref{fig:april_workflow} 
illustrates this stage, and
Alg.\ \ref{alg:stage1} presents the \ChM[pseudo code
for this stage]{corresponding pseudo code}. 
Below, we detail these four components.

\begin{algorithm}[t]
\SetKwInOut{Input}{Input}
\SetKwInOut{Output}{Output}
  \Input{Query budget $N$; document cluster $x$; Summary DB $D_S(x)$;
  heuristic $h$; tradeoff $\beta$, learning rate $\alpha$}
  let $\hat{U}_x = h$ \;
  get first summary $y_{1,1}$ by Eq. \eqref{eq:query} \;
  initialise $w_0$
  \While{$i=1, \ldots, N$}{
    select $y_{i,2}$ according to Eq. \eqref{eq:query} \;
    get preference $p_x^i(y_{i,1}, y_{i,2})$ from the oracle, add to $D_P$ \; 
    $w_i := w_{i-1} + \alpha \nabla_w \mathcal{R}^{\mathrm{BT}}_x(w)$ (Eq.~\eqref{eq:mal_bt}) \;
    $y_{i+1,1} = y_{i,2}$
    }
  $\Hat{U}_x(y) = (1-\beta) \cdot h(y,x) + \beta \cdot w_i \cdot \phi(y,x)$
    (Eq. \eqref{eq:hat_u}) \;
  \Output{$\hat{U}_x$ and its induced ranking $\Hat{\sigma}_x$}
  \caption{Active preference learning (Stage 1 in APRIL).
  \label{alg:stage1}}
\end{algorithm}

\paragraph{Summary DB $D_S(x)$.}
Ideally $D_S(x)$ should include all legal 
extractive summaries for a document cluster $x$,
namely $D_S(x) = \mathcal{Y}_x$.
Since this is impractical for large clusters, we either randomly sample a 
large set of summary candidates or use pre-trained summarization 
models and heuristics to generate $D_S(x)$.
Note that $D_S(x)$ can be built offline, i.e.\ before 
the interaction with the user starts.
This improves the
real-time responsiveness of the system.

\paragraph{Preference DB $D_P$.}
\ChM{The preference database stores all collected user preferences}
$D_P = \{p_x(\langle y_{1,1},y_{1,2} \rangle), \cdots, 
p_x(\langle s_{N,1},s_{N,2} \rangle)\}$,
where $N$ is the \emph{query budget}\ChM{\ (i.e., how 
many times a user may be asked for a preference)},
$y_{i,1}, y_{i,2} \in D_S(x)$ are the summaries presented to
the user in the $i^\mathrm{th}$ round of interaction, and
$p_x$ is the \ChM[oracle]{user}'s preference 
(see \S\ref{subsec:pref_learn}). 

\paragraph{Preference Learner.} 
We use the BT model introduced in \S\ref{subsec:pref_learn} to 
learn $\Hat{\sigma}_x$ from the preferences in $D_P$.
In order to increase \ChM{the} real-time responsiveness of our system,
we use a linear model to approximate the utility function $U^*_x$,
i.e. $\Hat{U}_x(y) = w \cdot \phi(y,x)$, where $\phi(y,x)$ is a
vectorised representation of summary $y$ for input cluster $x$.
However, purely using $w \cdot \phi(y,x)$ to approximate $U^*_x$
is sensitive to the noise in the preferences,
especially when the number of collected preferences
is small. To mitigate this,
we \ChM[estimate]{approximate} $U^*_x$ not only using 
$w \cdot \phi(s|x)$ (the posterior),
but also using some prior knowledge $h(y,x)$ (the prior),
for example
the heuristics-based summary evaluation function 
proposed by \citet{ryang2012emnlp}
and \citet{rioux2014emnlp}\ChM[ (n]{. N}ote that these heuristics do not
require reference summaries; see \S\ref{sec:related_work}\ChM[)]{}.
Formally, we define the $\Hat{U}_x$ as
\begin{align}
\hat{U}_x(y) = (1-\beta) \cdot h(y,x) + \beta \cdot w \cdot \phi(y,x),
\label{eq:hat_u}
\end{align}
where $\beta \in [0,1]$ is a real-valued parameter trading off between
the prior and posterior. 

\paragraph{AL-based Querier.}
\ChM{The active learning based querier}
receives $\hat{U}_x$ and selects which candidate \ChM{pair}
from $D_S$ to present to the user in each round of interaction.
To reduce the reading burden of the oracle,
inspired by the preference collection workflows in robots training
\citep{wirth_etal2016aaai_pbrl},
we use the following \ChM[querying workflow]{setup to obtain summary pairs}:
\ChM[in each]{In each interaction} round, one summary of the pair is
\emph{old} (i.e. it has been presented to the user in
the previous round) and the other one is \emph{new} 
(i.e.\ it has not been read by the user before). 
As such, the user only needs to read $N+1$ summaries
in $N$ rounds of interaction.

Any pool-based active learning strategy \citep{settles2010active_learning} 
can be used to implement the querier, e.g., uncertainty sampling \citep{lewis1994sequential}.
We explore four computationally efficient 
active learning strategies:
\begin{itemize}
  \item 
    \emph{Utility gap ($\Delta_{\hat{U}_x}$)}: 
    Inspired by the policy \ChM[in]{of} SPPI (see \S\ref{subsec:sppi} and Eq.~\eqref{eq:sr_query}),
    this strategy presents summaries
    with large estimated utility gaps $\Delta_{\hat{U}_x}$:
    \[\Delta_{\hat{U}_x}(y_i,y_j) = \hat{U}_x(y_i)-\hat{U}_x(y_j).\]
    
  \item \emph{Diversity-based heuristic $(div)$}: 
    This strategy minimises the vector space similarity
    of the presented summaries.
    For a pair $y_i, y_j \in \mathcal{Y}_x$, we define 
    \[div(y_i,y_j|x) = 1-\cos(\phi(y_i,x),\phi(y_j.x)),\]
    where 
    $\cos$ is the cosine similarity.
    This heuristic encourages querying dissimilar summaries,
    so as to encourage exploration and facilitate generalisation.
    In addition, dissimilar summaries are more likely
    to have large utility gaps\ChM[, hence are easier to answer]{\ 
    and hence can be answered more accurately by the users (discussed later
    in \S\ref{sec:pref_summary})}. 
    
  \item \emph{Density-based heuristic $(den)$}: 
    This strategy encourages querying summaries 
    from ``dense'' areas in the vector space, so as to 
    avoid querying outliers and to facilitate generalisation.
    Formally, for a summary $\ChM[s]{y}$ for cluster $x$, 
    we define 
    \[den(y|x) = 1 - \min_{y' \in D_S(x), 
    y' \not = y} div(y, y'|x).\]
  \item 
    \emph{Uncertainty-based heuristic $(unc)$}: 
    This strategy encourages querying the summaries
    whose approximated utility $\hat{U}_x$ is most uncertain.
    In line with \citet{avinesh_meyer2017_interactive_doc_sum},
    we define $unc$ as follows: For \ChM{a} summary $y \in D_S(x)$, 
    we estimate the probability of $y$ being the optimal summary as  
    \[pb(y|x) = \frac{1}{1+\exp[-\hat{U}_x(y)]}\,,\]
    and let the uncertainty of $y$ be $unc(y|x) = 1-pb(y|x)$ 
    if $pb(s|x) \ge 0.5$, and let $unc(y|x) = pb(y|x)$ otherwise.
\end{itemize}

\noindent
To exploit the strengths of all these AL strategies, 
we normalise their output values to the same range
and use their weighted sum to select the 
new summary $y^*$ to present to the user:
\begin{align}
  y^* = & \arg\max_{y \in D_S(x)} \: 
  [\:  w_g \cdot |\Delta_{\hat{U}_x}(y,y')| + w_d \cdot div(y,y'|x) 
    \nonumber \\
  & \hspace*{2cm} +  w_e \cdot den(y|x) + w_u \cdot unc(y|x) \,],
  \label{eq:query}
\end{align}
where $y'$ is the old summary,
i.e.\ the one from the previous interaction round.
To select the first summary, we let $div(y,y')=0$
and $\Delta_{\hat{U}_x}(y,y')=\hat{U}_x(y)$.
$w_g$, $w_d$, $w_e$ and $w_u$ denote the weights
for the four heuristics.

\subsection{Stage 2: RL-based Summariser}
\label{subsec:stage2}
Given the approximated ranking $\Hat{\sigma}_x$ learnt by the first stage,
the target of the second stage in APRIL is to obtain 
\begin{align*}
\Hat{\pi}^* = \arg \max_{\pi} \hat{\mathcal{R}}^{\mathrm{APRIL}}_x(\pi)
= \arg \max_{\pi} \sum_{y \in \mathcal{Y}_x(\pi)} \pi(y) \Hat{\sigma}_x(y).
\end{align*}
We consider two RL algorithms to obtain $\Hat{\pi}^*$:
the linear \emph{Temporal Difference} (TD) algorithm,
and \ChM[the neural-based TD algorithm]{a neural version of the TD algorithm}.

\ChM[The \emph{Temporal Difference} (TD) algorithm]{TD}
\citep{Sutton84} has proven effective for solving the MDP in EMDS
\citep{rioux2014emnlp,ryang2012emnlp}.
The core of TD is to approximate the \emph{$V$-values}:
In EMDS, $V^\pi(s)$ estimates the ``potential''
of the (draft) summary $s$ for input cluster $x$ given policy $\pi$:
the higher the $V^\pi(s)$ value, the more likely 
$s$ is contained in the optimal summary for $x$.
Given the $V$-values, a policy can be derived  using
the \emph{softmax} strategy:
\begin{align}
\pi(s,a) = \frac{\exp[V^\pi(P(s,a))]}{\sum_{a'}\exp[V^\pi(P(s,a'))]},
\label{eq:softmax-rl}
\end{align}
where $a'$ ranges over all available actions in the state $s$.
The intuition behind Eq.\ \eqref{eq:softmax-rl} is that
the probability of performing the action $a$ increases if the resulting
state of $a$, namely $P(s,a)$, has a higher $V$-value.
Note the similarity between the policy of TD (Eq.~\eqref{eq:softmax-rl})
and the policy of SPPI (Eq.~\eqref{eq:sr_query}):
they both use a Gibbs distribution to assign probabilities to 
different actions, 
but the difference is that an action in SPPI
is a pair of summar\ChM[y]{ies}, while in TD an action is adding a sentence
to the current draft summary or \emph{terminate} (see \S\ref{subsec:rl}).

Existing works use linear functions to approximate the $V$-values
\citep{rioux2014emnlp,ryang2012emnlp}.
To more precisely approximate the $V$-values,
we use a neural network
and term the resulting algorithm \emph{Neural TD} (NTD). 
Inspired by DQN \citep{mnih2015human}, 
we employ the \emph{memory replay} 
and \emph{periodic update} techniques
to boost and stabilise the performance of NTD.
We use NTD rather than DQN \citep{mnih2015human}
because in MDPs with discrete actions and continuous states,
as in our EMDS formulation, Q-Learning needs to maintain a $Q(s,a)$
network for each action $a$, which is very expensive when
the size of $a$ is large. Instead, the TD algorithms only
have to maintain the $V(s)$ network, whose size is independent of
the number of actions. In EMDS, the size of 
the action set typically exceeds several hundreds
(see Table \ref{table:datasets}), 
because each sentence corresponds to one action.

Alg.\ \ref{alg:deep_td} shows the pseudo code of NTD.
We use the Summary DB $D_S$ as the memory replay.
This helps us to reduce the sample generation time,
which is critical in interactive systems.
We select samples from $D_S$ using softmax sampling
(line 3 in Alg.\ \ref{alg:deep_td}):
\begin{align}
\mathcal{P}(y;\theta|x) = \frac{\exp[V(y;\theta)]}{\sum_{y'\in D_S(x)}\exp[V(y';\theta)]},
\label{eq:imp_sample}
\end{align}
where $\theta$ stands for the parameters of the neural network.
Given the selected summary $y$, we build a sequence of $k$ states
(line 4), where $k$ is the number of sentences in $\ChM[s]{y}$, and state $s_i$
is the draft summary including the first $i$ sentences
of $\ChM[s]{y}$. Then, we update the \ChM[TD error]{error} $\mathcal{L}^\mathrm{TD}$
as in the standard TD algorithms (lines 5 to 9)
and perform back propagation with gradient descent (line 10).
We update $\theta'$ every $C$ episodes (line 11), as
in DQN, to stabilise the performance of NTD.
After finishing all training,
the obtained $V$-values can be used to derive the $\Hat{\pi}^*$
by Eq.~\eqref{eq:softmax-rl}.

\begin{algorithm}[t]
\SetKwInOut{Input}{Input}
\SetKwInOut{Output}{Output}
  \Input{
  Learning episode budget $T$; 
  document cluster $x$;
  summary DB $D_S(x)$; approximated ranking function $\hat{\sigma}_x$; 
   update frequency $C$}
  \While{$t=1, \ldots, T$}{
    initialise $\theta$ randomly, let $\theta' = \theta$\;
    sample $y \in D_S(x)$ by Eq.\ \eqref{eq:imp_sample}\;
    build states from $y$: $s_1, \ldots, s_k$\;
    $\mathcal{L}^\mathrm{TD} = 0$\;
    \While{$i = 0, \ldots, k-1$}{
      $r_i =  \left\{ \begin{array}{l l}
      \hat{\sigma}_x(y) & \text{if } i+1 = k \\
      V(s_{i+1},x;\theta') & \mbox{otherwise}
      \end{array} \right.$ \;
      $\mathcal{L}^\mathrm{TD} = \mathcal{L}^\mathrm{TD} + (r_i - V(s_i,x;\theta))^2$
    }
    update $\theta$ with gradient descent\;
    let $\theta'=\theta$ every $C$ episodes\;
    }
  \Output{Policy $\hat{\pi}^*$ by Eq.~\eqref{eq:softmax-rl}}
  \caption{NTD algorithm for EMDS.
  \label{alg:deep_td}}
\end{algorithm}


\section{Preference-based Interaction for Summarisation}
\label{sec:pref_summary}

To date, there is little knowledge about the usability and the reliability of user feedback in summarisation.
This is a major limitation for designing interactive systems and for effectively experimenting with simulated users before an actual user study.
In this section, we therefore study preference-based feedback for our EMDS use case and derive a mathematical model to simulate real users' preference-giving behaviour.

\paragraph{\ChM{Hypotheses}.}
Our study tests two hypotheses:
(\ChM[i]{H1})~We assume that users find it \ChM[easy]{easier} to provide preference feedback \ChM{than \ChM[using]{providing} 
other forms of feedback for summaries}. 
In particular, \ChM[in \S\ref{subsec:user_friendly_study}, w]{w}e measure the user satisfaction and the time needed for 
preference-based interaction and bigram-based interaction
proposed by \citet{avinesh_meyer2017_interactive_doc_sum},
which has also been used in interactive summarisation.

(\ChM[ii]{H2})~\ChM{Previous research suggests that the more difficult the questions, the lower the correct rate of the answers or, in other words, the higher the noise in the answers
\citep{DBLP:conf/nips/HuangLVA016,DBLP:conf/cikm/DonmezC08}.
In our preference-giving scenario, we assume that the difficulty
of comparing a pair of items 
can be measured by the \emph{utility gap} between the 
presented items:
the wider the utility gap, the easier it is for the 
user to identify the better item.
We term this the \emph{wider-gap-less-noise} hypothesis
in this article.
}
%

\ChM{The wider-gap-less-noise hypothesis is an essential motivation 
for \YG[APRIL's APL design]{ the policy in SPPI (Eq.~\eqref{eq:sr_query})
and the diversity-based active learning strategy in APRIL 
(see \S\ref{subsec:stage1})}, but yet there is little empirical evidence for validating this hypothesis.
Based on the findings in our user study, we provide evidence 
towards H1 and H2, and we propose a realistic user model, 
which we employ in our simulation experiments in \S\ref{sec:experiments}.
}

\paragraph{Study setup.}
We invite 12 users to participate in our user study. 
All users are native or fluent English speakers from our university.
We ask each user to provide feedback for newswire summaries from two topics (d074b from DUC'02 and d32f from DUC'01) in the following way.

We first allow the users to familiarise with the topic by means of two 200-words abstracts.
This is necessary, since the events discussed in the news documents are several years old and maybe unknown to our participants.
Without having such background information, it would not be possible for users to judge importance in the early stages of the study.
We ask each user to provide preferences for ten summary pairs and to label all important bigrams in five additional summaries.
For collecting preference-based feedback, we ask the participants to select the better summary 
(i.e. the one containing more important information) in each pair. 
For collecting bigram-based feedback, we adopt the setup of \citet{avinesh_meyer2017_interactive_doc_sum}, who proposed a successful EMDS system using bigram-based interaction.
At the end of the study, we ask the participants to rate the usability (i.e., user-friendliness)
of preference- and bigram-based interaction on a 5-point Likert scale,
where higher scores indicate higher usability.

To evaluate H2, we require summary pair with different utility gaps.
To this end, we measure the utility $U^*_x$ 
(see \S\ref{subsec:pref_learn}) of a summary $y$ for document cluster $x$ as
\begin{align}
U^*_x(y) = \ChM[3.3\times]{\frac{10}{3}} \left(\frac{R_1(y,r_x)}{0.47} + \frac{R_2(y,r_x)}{0.22} + 
\frac{R_\mathrm{SU}(y,r_x)}{0.18}\right),
\label{eq:u_star}
\end{align}
\YG{where $r_x$ are the reference summaries for document cluster $x$
(provided in the DUC datasets), and
$R_1$, $R_2$ and $R_\mathrm{SU}$ stand for average 
ROUGE-1, ROUGE-2 and ROUGE-SU4 recall metrics \citep{lin2004rouge},
respectively.}  These ROUGE metrics are widely used
to measure the quality of summaries. 
The denominator values 0.47, 0.22 and 0.18 are the 
upper-bound ROUGE scores reported by \citet{avinesh_meyer2017_interactive_doc_sum}. 
They are used to balance the weights of the three ROUGE scores.
As such, each ROUGE score is normalised to $[0,1]$, and
we further multiply the sum of the ROUGE scores by \ChM[3.3]{$\frac{10}{3}$} to normalise $U^*_x$ values 
to $[0,10]$, which facilitates our analyses afterwards.

For document cluster $x$,
the utility gap $\Delta_{U^*_x}$ of two summaries $s_1$ and $s_2$
is thus $\Delta_{U^*_x}(y_1,y_2) = U^*_x(y_1)-U^*_x(y_2)$.
As for the ten summary pairs in our user study, 
we select four pairs with utility gap $\tilde{\Delta} = 1$, 
three with $\tilde{\Delta} = 3$, two with $\tilde{\Delta} 
= 5$ and one with $\tilde{\Delta} = 7$, 
where $\tilde{\Delta} = |\Delta_{U^*_x}| \pm .1$ 
(i.e., a utility gap very close to the predefined gap width). 
Figure \ref{fig:example_pairs} shows 
two example summary pairs and their $U^*_x$.
As for the five summaries for bigram-based feedback, 
we select summaries with high utility $U^*_x$, 
but ensure that they have low overlap in order simulate the setup AL setup of \citet{avinesh_meyer2017_interactive_doc_sum}.

\begin{figure*}
\newcommand{\Summary}[2]{\framebox{%
  \begin{minipage}[t][4.2cm][t]{.47\linewidth}%
    \textbf{#1}: {#2}%
  \end{minipage}%
}}
\scriptsize
\centering
\Summary{$\mathbf{Summ}_{A,1} (U^* = 3.99)$}{
``I think he's doing a beautiful job up there. ``; President Bush, asked at a news conference whether Thomas' claim not to have an opinion on abortion is credible, answered, ``That's a question for the Senate to decide. In their respective careers, the Thomases have embraced the view that women and minorities are hindered, rather than helped, by affirmative action and government programs. True equality is achieved by holding everyone to the same standard, they believe. ``; Before Thomas' testimony ended, the unflappable 43-year-old federal judge was criticized, sometimes in harsh terms, by several liberal Democrats. Hatch asked.
}
\Summary{$\mathbf{Summ}_{A,2} (U^* = 5.02)$}{
They see a woman with strong opinions on issues that are bound to come before the court. Dean Kelley, the National Council of Churches' counselor on religious liberty, wrote a critique of Clarence Thomas that was used as grounds for his organization's opposition to the Supreme Court nominee. ``; Thomas said Senate confirmation of his nomination would give him ``an opportunity to serve and give back'' and to ``bring something different to the court. True equality is achieved by holding everyone to the same standard, they believe. ``; ``He's handling himself very well,'' the president said. Hatch asked.
}
\Summary{$\mathbf{Summ}_{B,1} (U^* = 6.52)$}{
Heflin cited the ``appearance of a confirmation conversion'' and said it may raise questions of Thomas' ``integrity and temperament. The ministers were recently organized into a conservative Coalition for the Restoration of the Black Family and Society, with the first item on its agenda being Thomas' confirmation. After still another Thomas answer, Biden said, ``That's not the question I asked you, judge. Several committee members said they expected the committee to recommend, by a 10-4 or 9-5 vote, that the Senate confirm Thomas. But others see a different symbolism. But others see a different symbolism. But they hope Sens.
}
\Summary{$\mathbf{Summ}_{B,2} (U^* = 1.46)$}{
During the early '80s, Virginia Thomas enrolled in Lifespring, a self-help course that challenges students to take responsibility for their lives. RADIO; (box) KQED, 88.5 FM Tape delay beginning at 9 a.m. repeated at 9:30 p.m. (box) KPFA, 94.1 FM Live coverage begins at 6:30 a.m. TELEVISION; (box) C-SPAN Live coverage begins at 7 a.m. repeated at 5 p.m. (box) CNN Intermittent coverage. ``; On natural law : ``At no time did I feel, nor do I feel now, that natural law is anything more than the background to our Constitution. ``I'm not satisfied with the answers,'' Leahy said.
}
\caption{Two summary pairs from topic d074b with utility gaps $\tilde\Delta = 1$
(pair A, the upper two summaries) and $\tilde\Delta = 5$ (pair B, the bottom two summaries).}
\label{fig:example_pairs}
\end{figure*}



\paragraph{Usability assessment.}
\label{subsec:user_friendly_study}
\ChM[W]{To evaluate hypothesis H1, w}e measure the easiness of providing preferences
for summaries with two metrics: 
the average \ChM[time]{\emph{interaction time}} a \ChM[user]{participant} spends in providing a preference 
and \ChM[users' rating of the usability]{the participant's \emph{usability rating} on the 5-point scale}.
\ChM{We compare both metrics for preference-based interaction with bigram-based interaction.}

\begin{figure}
  \begin{subfigure}[b]{0.45\textwidth}
    \includegraphics[width=\textwidth]{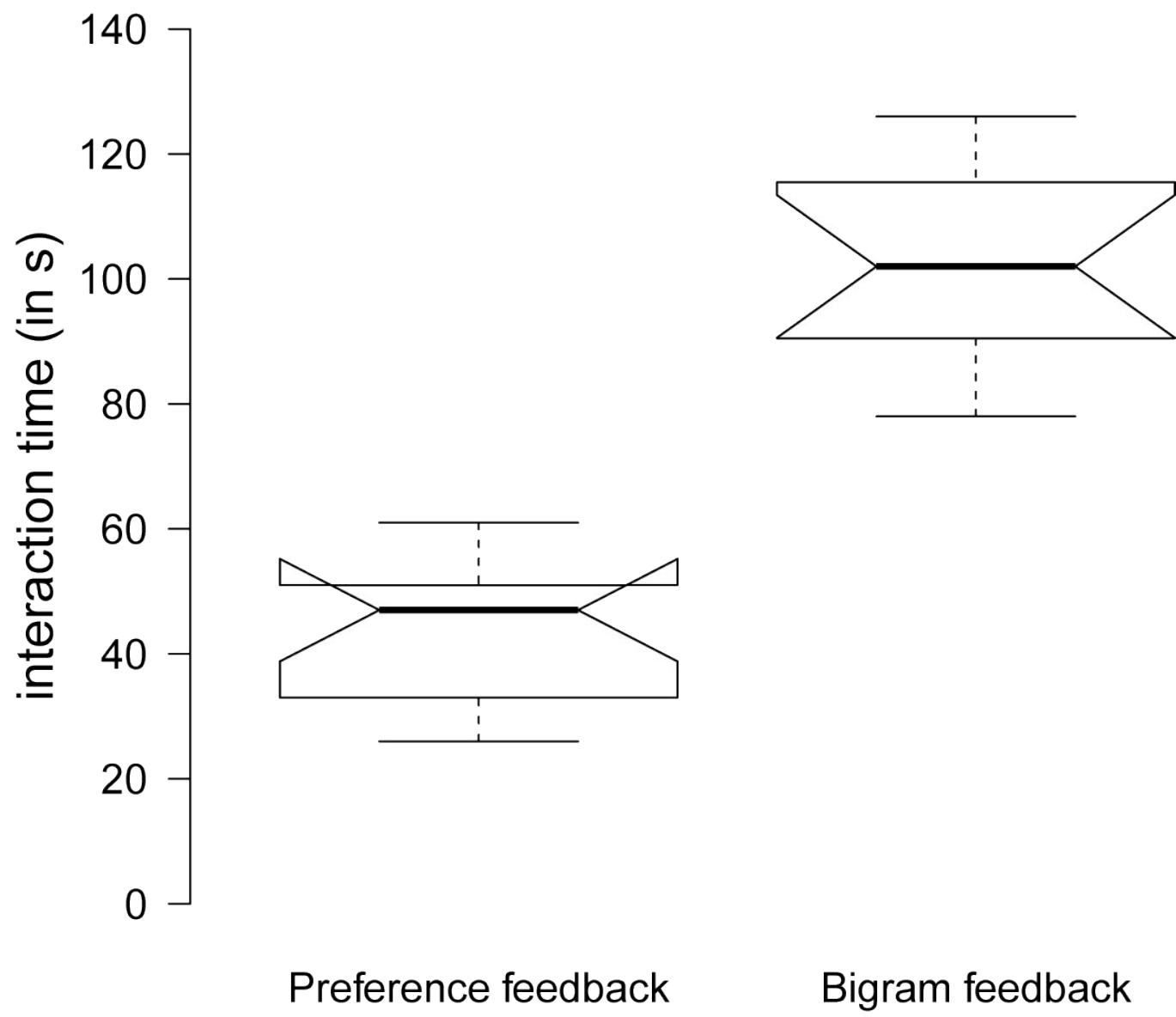}
    \caption{Interaction time} 
  \end{subfigure}
  \qquad\quad
  \begin{subfigure}[b]{0.45\textwidth}
    \includegraphics[width=\textwidth]{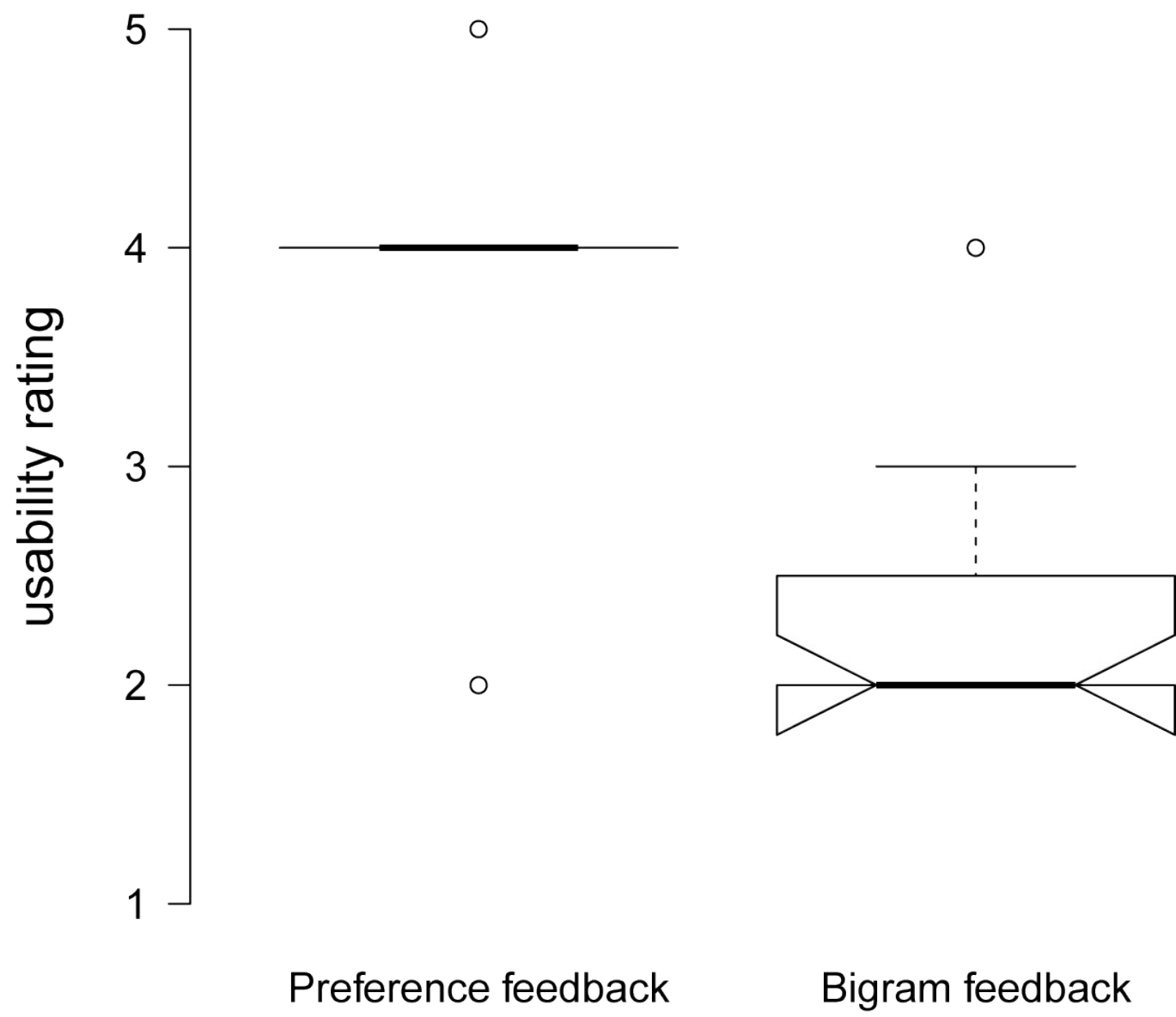}  
    \caption{Usability ratings}
  \end{subfigure}
  \caption{Comparison of interaction time and usability ratings for preference and bigram-based interaction. 
  Notches indicate 95\% confidence interval.
  }
  \label{fig:userstudy1}
\end{figure}

Fig.~\ref{fig:userstudy1} visualises the interaction time and the usability ratings for preference and bigram-based interaction as \ChM[a bar chart and 
a notched boxplot, respectively]{notched boxplots}. Both plots confirm the clear difference between preference- and bigram-based feedback for summaries:
We measure an average \ChM[processing]{interaction} time of 
\ChM[101]{102}\,s (with standard error $\mathit{SE} = \ChM[6]{4\,s}$) 
for annotating bigrams in a \ChM[document]{single summary},
which is over twice the time spent \ChM[on]{for} providing a preference
for a summary pair (43\,s, $\mathit{SE} = \ChM[2]{3\,s}$).
The users identified 7.2 bigrams per summary, which
took 14s per bigram on average.
As for the \ChM[user-friendliness]{usability} ratings, 
providing preferences is rated 
3.8 ($\mathit{SE} = 0.\ChM[49]{27}$) on average\ChM{\ (median at 4)}, 
while labelling bigrams is rated 2.4 ($\mathit{SE} = 0.\ChM[4]{22}$) on average\ChM{\ (median at 2)}. 
These results suggest that \ChM[people]{humans} can more easily
provide preferences over summaries than providing 
\ChM[other forms of feedback, for example the bigrams]{point-based feedback in the form of bigrams}.

\paragraph{Reliability assessment.}
\label{subsec:noise_in_pref}

\begin{figure}[t]
  \centering
  \includegraphics[trim={0 0 0 0.7cm},clip,width=0.7\textwidth]{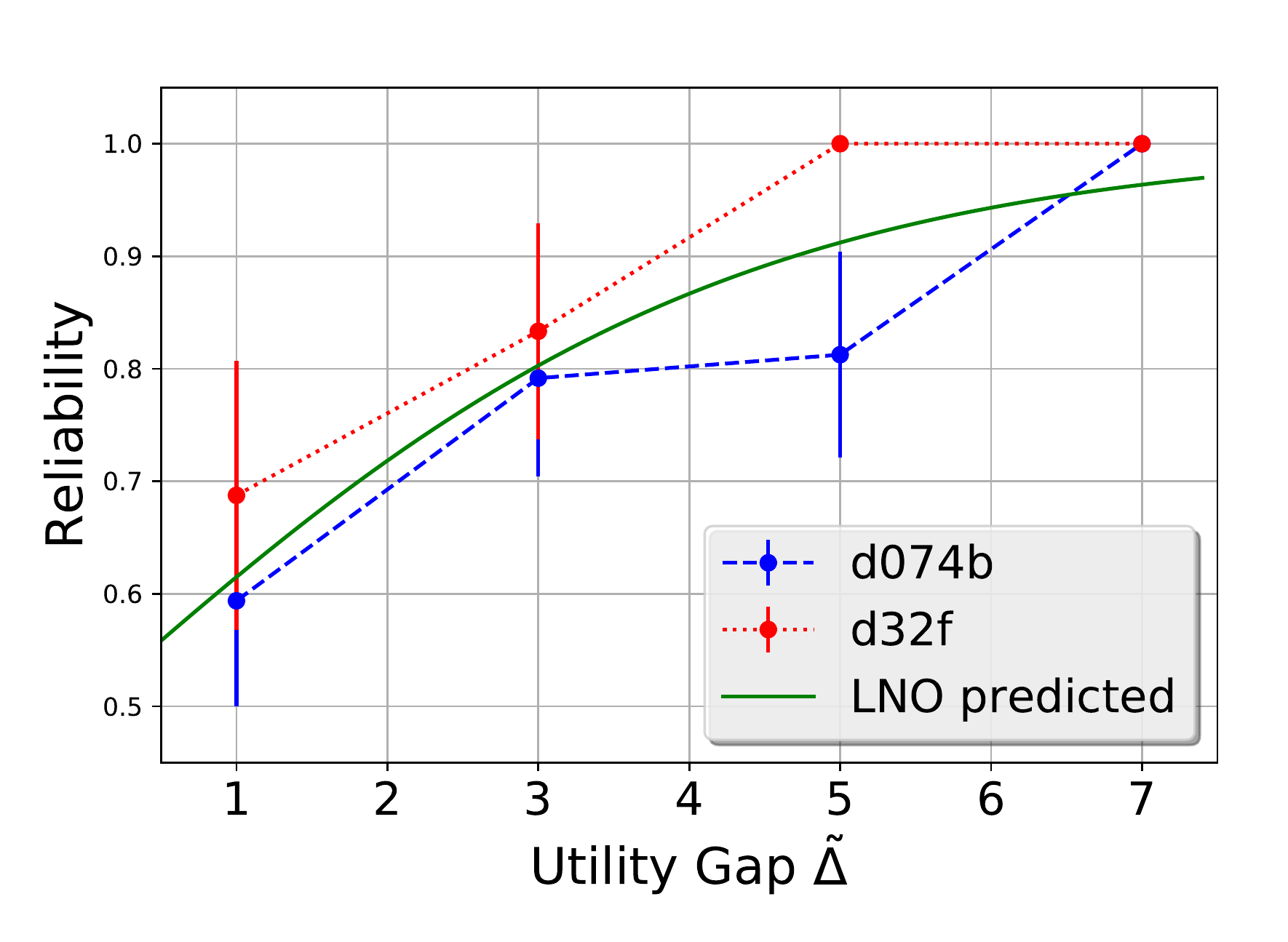}
  \caption{The reliability of users' preferences 
  increases with the growth of the utility gaps
  between presented summaries. 
  Error bars indicate standard errors.
 }
  \label{fig:acc_line}
\end{figure}

\ChM[Figure \ref{fig:acc_line} presents the \emph{accuracy}]{To evaluate hypothesis H2, we measure the \emph{reliability}}
of the users' preferences, i.e. the percentage of the pairs 
in which the user's preference is the
same as the preference induced by $U^*$.
\ChM[It]{Figure~\ref{fig:acc_line} shows the reliability scores for the varying utility gaps employed in our study. The results}
clearly suggest that, for summary pairs with
wider utility gaps, \ChM[users]{the participants} can more easily identify
the better summary in the pair, resulting
into higher reliability. This observation validates the
wider-gap-less-noise assumption.

\paragraph{\ChM{Realistic user simulation.}}
\ChM[Note]{We observe} that the shape of the \ChM[accuracy]{reliability} curves in 
Figure \ref{fig:acc_line} is similar to that of the
logistic function: when $\tilde{\Delta}$ approaches
0, \ChM[accuracy]{the reliability scores} approaches 0.5\ChM[;]{\ and} with the increase of 
$\tilde{\Delta}$, the \ChM[accuracy]{reliability} asymptotically approaches 1. 
Hence, we \YG{adopt the} logistic model proposed by
\citet{DBLP:conf/nips/ViappianiB10}
to estimate the real users' preferences.
We term the model \emph{logistic noise oracle} (LNO):
For two summaries $y_i,y_j \in \mathcal{Y}_x$,
we assume the probability that \ChM[the]{a} user 
prefers $y_i$ over $y_j$ is:
\begin{align}
\mathcal{P}_x(y_i\succ y_j;m) = \left(
1 + \exp\left[\frac{\Delta_{U^*_x}(y_j)-\Delta_{U^*_x}(y_i)}{m}\right]
\right)^{-1}\,,
\label{eq:lno}
\end{align}
where $m$ is a real-valued parameter controlling 
the ``flatness'' of the curve: higher $m$ \ChM[results in]{yield a}
flatter curve, which in turn suggests \ChM[the user has high noise in distinguishing]{that asking users to distinguish}
summaries with 
similar quality\ChM{\ causes high noise}.

We estimate $m$ based on the observations we made
in the user study by maximising the likelihood function:
\begin{align*}
l^\mathrm{LNO}(m) = & \sum_u \sum_n [
p_u(\langle y_{i,1}, y_{i,2} \rangle) 
\log \mathcal{P}_x(y_{i,1} \succ y_{i,2};m)
\nonumber \\& 
+ p_u(\langle y_{i,2}, y_{i,1} \rangle) 
\log \mathcal{P}_x(y_{i,1} \prec y_{i,2};m)]
\end{align*}
where $u$ ranges over all users and $i$
ranges over the number of preferences provided by each user.
$y_{i,1}$ and $y_{i,2}$ are the summaries presented to the user in round $n$.
$p_u$ is the user's preference direction function:
$p_u(\langle y_{i,1}, y_{i,2} \rangle)$ equals 1 if $y_{i,1}$ is
preferred by the user over $y_{i,2}$, and equals 0 otherwise.
By letting $\ChM[\partial \mathcal{L}^\mathrm{LNO}(m)/\partial m]{\frac{\partial}{\partial m} \mathcal{L}^\mathrm{LNO} (m)} = 0$,
we obtain $m=2.14$. 
The green curve in Figure \ref{fig:acc_line} is the 
\ChM[accuracy]{reliability} curve for the LNO with $m=2.14$. 
We find \ChM{that} it fits well
with the \ChM[accuracy]{reliability} curves of the real users.
As a concrete example, consider the
summary pairs in Figure \ref{fig:example_pairs}:
LNO prefers $\mathbf{Summ}_{A,2}$ over $\mathbf{Summ}_{A,1}$ with 
probability $.618$ and prefers $\mathbf{Summ}_{B,1}$ over $\mathbf{Summ}_{B,2}$ 
with probability  $.914$, \ChM{which is }consistent with our observations that 
7 out of 12 users prefer $\mathbf{Summ}_{A,2}$ over $\mathbf{Summ}_{A,1}$, 
while all users prefer $\mathbf{Summ}_{B,1}$ over $\mathbf{Summ}_{B,2}$.




\section{Simulation Experiments}
\label{sec:experiments}
\YG{In this section, we study APRIL in a simulation setup.}
We use the LNO-based user model with $m=2.14$ to simulate
user preferences \ChM[(see \S\ref{sec:pref_summary_study})]{as introduced in \S\ref{sec:pref_summary}}.
We separately study the first and \ChM{the} second stage 
\ChM[in]{of} APRIL,
by comparing multiple \YG[APL]{active learning} 
and RL techniques in each stage.
Then, we combine the best-performing strategy from 
each stage to build the overall \ChM{APRIL} pipeline and
compare our method with SPPI. 
\YG{We perform our experiments
on three multi-document summarisation benchmark
datasets from  the Document Understanding 
Conferences\footnote{\url{https://duc.nist.gov/}} (DUC)}:
DUC'01, DUC'02 and DUC'04. Table~\ref{table:datasets} shows 
the main properties of these datasets.
To ease the reading, we summarise the parameters we used
in our simulation experiments in Table~\ref{table:parameters}.

\begin{table}[h]
  \centering
  \newcolumntype{S}{@{\hspace{3mm}}}
    \begin{tabular}{l c S c S c S c}
      \toprule
      Dataset & \#\,Topic & \#\,Doc & SumLen & \#\,Sent/Topic \\
      \midrule
      DUC$\,$'01 &  30 & 308 & 100 & 378 \\
      DUC$\,$'02 &  59 & 567 & 100 & 271\\
      DUC$\,$'04 &  50 & 500 & 100 & 265\\
      \bottomrule
    \end{tabular}
  \caption{\ChM[Our simulation experiments were performed
  on three DUC datasets. ]{For our experiments, we use standard benchmark datasets from the Document Understanding Conference (DUC).}  
  \#\,Doc: the overall number of documents across all topics.
  SumLen: the length of each summary (in tokens).
  \#\,Sent/Topic: average number of sentences in a topic.}
  \label{table:datasets}
\end{table}

\subsection{APL Strategy Comparison}
\label{subsec:query_strategy_compare}

%
We compare our AL-based querying strategy 
introduced in \S\ref{subsec:stage1}
(see Eq.\ \eqref{eq:query}) 
with three baseline AL strategies:
\begin{itemize}
\item \emph{Random}:
In each interaction round, select a new 
candidate summary from $D_S$ 
uniformly at random and ask the user to compare it to the old 
one from the previous interaction round. 
In the first round, we randomly select two summaries to present.
\item \emph{J\&N}
is the \emph{robust query selection algorithm}
proposed by \cite{DBLP:conf/nips/JamiesonN11}.
It assumes \ChM{that the }items' preferences are dependent on
their distances to an unknown reference point in the 
embedding space: the farther an item to the reference
point, the more preferred the item is. After each round
of interaction, the algorithm uses all collected preferences
to locate the area where the reference point may fall into and identif\ChM[y]{ies} the query pairs which can reduce the size of this
area, termed \emph{ambiguous query pairs}. 
To combat noise in preferences, the algorithm selects the 
most-likely-correct ambiguous pair
to query the oracle in each round.
\item \emph{Gibbs}:
This is the querying strategy used in SPPI.
In each round, it selects summaries $y_i,y_j$ with 
the Gibbs distribution (Eq.~\ref{eq:sr_query}),
and updates the weights for the utility function
as in line 5 in Alg. \ref{alg:sr}. 
\end{itemize}
Note that Gibbs presents two new summaries
to the user each round, while the
other querying strategies we \ChM[considered]{consider}
present only one new summary per round (see
\S\ref{subsec:stage1}). 
Thus, \ChM[with]{in} $N$ rounds of interaction with a user, 
the user needs to read $2N$ summaries
with Gibbs, but only $N+1$ with the other querying strategies.

\begin{table}[t]
    \centering
    \begin{tabular}{l p{.7\linewidth}}
    \toprule
    Parameter & Description \\
    \midrule
    \multicolumn{2}{l}{\textbf{For APL (stage 1 in APRIL); see Alg.~\ref{alg:stage1}:}} \\
    $N = 10, 50, 100$ & query round budget  \\
    $|D_S(x)| = 5000$ & Summary DB size for each cluster $x$ (see \S\ref{subsec:stage1}) \\
    $h$ & heuristics-based prior reward (see \S\ref{subsec:stage1}
    and Eq.~\eqref{eq:hat_u}); we use the reward heuristics proposed
    by \citet{ryang2012emnlp}\\
    $\beta = 0.5$ & trade-off between prior and posterior rewards
    (see Eq.~\eqref{eq:hat_u}) \\
    $\alpha = 10^{-3}$ & learning rate for preference learning \\
    $\phi(y,x)$ & vectorised representation of summary $y$ for document
    cluster $x$ (see Eq.~\eqref{eq:hat_u}); we use the same \ChM[vector 
    representation as in \citep{rioux2014emnlp}]{vector representation as \citet{rioux2014emnlp}} \\
    $w_d=1$ & weights of the preference learning strategies (see Eq.~\eqref{eq:query};
    selection details presented in \S\ref{subsec:query_strategy_compare}) \\
    \multicolumn{2}{l}{\textbf{For RL (stage 2 in APRIL); see Alg.~\ref{alg:deep_td}:}} \\
    $T=3000$ & episode budget \\ 
    $C=50$ & update frequency in NTD \\
    $V(s,x;\theta)$ & neural approximation of $V$-values (see \S\ref{subsec:rl_comparison} for setup details) \\
    \multicolumn{2}{l}{\textbf{For SPPI; see Alg.~\ref{alg:sr}:}} \\
    $\gamma = 10^{-3}$ & learning rate in SPPI.\\
    \bottomrule
    \end{tabular}
    \caption{Overview of the parameters used in simulation experiments.}
    \label{table:parameters}
\end{table}

To find the best weights $w_g$, $w_d$,
$w_e$ and $w_u$ for our AL querying strateg\ChM[ies]{y} 
in Eq.\ \eqref{eq:query}, we run grid search:
We select each weight from $\{0,0.2,0.4,0.6,0.8,1\}$
and ensure that the sum of the four weights is 1.0.
The query budget $N$ was set to 10, 50 and 100.
For each cluster $x$, we generated 5,000 extractive
summaries to construct $D_S(x)$. 
Each summary contains no more than 100 words, 
generated by randomly selecting sentences
in the original documents in $x$.
The prior $h$ used in Eq. \ref{eq:hat_u}
is the reward function proposed by \citet{ryang2012emnlp}\ChM[ The]{, and we set the}
 trade-off parameter \ChM[$\beta=0.5$, as determined in pilot experiments]{$\beta$ to $0.5$}.
%
All querying strategies we test take less than 500\,ms
to decide the next summary pair to present.

The performance of the querying strategies is measured by 
the quality of their resulting reward function $\hat{U}_x$
(see Eq.~\eqref{eq:hat_u}).
For each cluster $x$,
we measure the quality of $\hat{U}_x$ by its Spearman's rank
correlation \citep{spearman1904proof} to the 
gold-standard utility scores $U^*_x$ (Eq.\ \eqref{eq:u_star})
over all summaries in $D_S(x)$.
We normalise $\hat{U}_x$ to the same range of $U^*_x$ (i.e. [0,10]).
For the vector representation $\phi$,
we use the same \ChM{200-dimensional} bag-of-bigram representation 
as \citet{rioux2014emnlp}. 

\begin{table}
  \centering\small
    \begin{tabular}{l l l l }
      \toprule
       & $N\!=\!10$ & $N\!=\!50$ & $N\!=\!100$ \\
      \midrule
      Random & .232 & .235 & .243 \\
      J\&N & .238 & .240 & .247 \\
      Gibbs & .246 & .275 & .289 \\
      \midrule
      $\Delta_{\hat{U}}$ ($w_g=1$) & .236 & .241 & .261 \\
      $div$ ($w_d=1$) & \textbf{.288$^*$} & .297$^*$ & .319$^*$ \\
      $den$ ($w_e=1$) & .211 & .238 & .263 \\
      $unc$ ($w_u=1$) & .257$^*$ & .285$^*$ & .303$^*$ \\
      BestCombination & \textbf{.288$^*$} & \textbf{.298$^*$} & \textbf{.320$^*$} \\
      \midrule 
      \multicolumn{4}{l}{Lower bound, $N=0$, $\beta=0$: .194} \\
      \bottomrule
    \end{tabular}
  \caption{Spearman's rank correlation between $\hat{U}_x$
  and $U^*_x$, averaged over 20 independent runs
  on all clusters $x$ in DUC'01. 
  $\hat{U}_x$ is learnt with different
  querying strategies. The lower bound is to 
  prohibit all interactions ($N=0$) and let $\hat{U} = h$
  (i.e. $\beta=0$ in Eq. \ref{eq:hat_u}).
  \ChM[Asterisk:]{Results marked with an asterisk are} 
  significantly better than all baselines.
  }
  \label{table:query_perf}
\end{table}

Table \ref{table:query_perf} compares the performance
of different querying strategies.
We find that all querying strategies outperform the
zero-interaction lower bound even with 10 rounds of interaction,
suggesting that even collecting a small number of preferences
can help to improve the quality of $\hat{U}_x$.
Among all baseline strategies, Gibbs significantly\footnote{
We used double-tailed t-tests to compute the $p$-values,
and selected $p<0.01$ as the significance level.}
outperforms the other two, and we believe the reason is that
Gibbs exploits the wider-gap-less-noise assumption\ChM{\ (see \S\ref{sec:pref_summary})}.
Of all 56 possible AL weights combinations,
48 combinations outperform the random
and J\&N baselines, 
and 27 outperform Gibbs. This \ChM[suggests]{shows} the 
overall strength of our AL-based strateg\ChM[ies]{y}.
The best combination of the weights
is $w_g=0$, $w_d=0.6$ and $w_e = w_u = 0.2$,
closely followed by using the
diversity-based strategy $div$ alone (i.e. $w_d=1$).
We believe the reason behind the effectiveness of
the $div$ strategy is that it not only exploits
the wider-gap-less-noise assumption by querying
\ChM[unsimilar]{dissimilar} summaries, but also explores summaries
from different areas in the embedding space, which 
helps the generalisation.
\ChM{Due to its simplicity, we use $w_d = 1$ henceforth, since its performance is almost identical
\YG{and has no statistically significant difference} 
to the best combination.}

\subsection{RL Comparison}
\label{subsec:rl_comparison}
We compare NTD (Alg.\ \ref{alg:deep_td}) to two baselines:
TD \citep{Sutton84} and
LSTD \citep{DBLP:conf/icml/Boyan99}. 
TD has been successfully used by
\citet{ryang2012emnlp} and \citet{rioux2014emnlp} 
for EMDS\ChM[;]{.} LSTD \ChM[is improved based on]{improves} TD
by using least square optimisation, and it has been \ChM[proved]{proven} to
\ChM[have better performance]{perform better} in large-scale problems
than TD \citep{lagoudakis2003lspi}.
Note that both TD and LSTD uses linear models to approximate the 
$V$-values.

\begin{figure}
\centering
\includegraphics[width=0.8\textwidth]{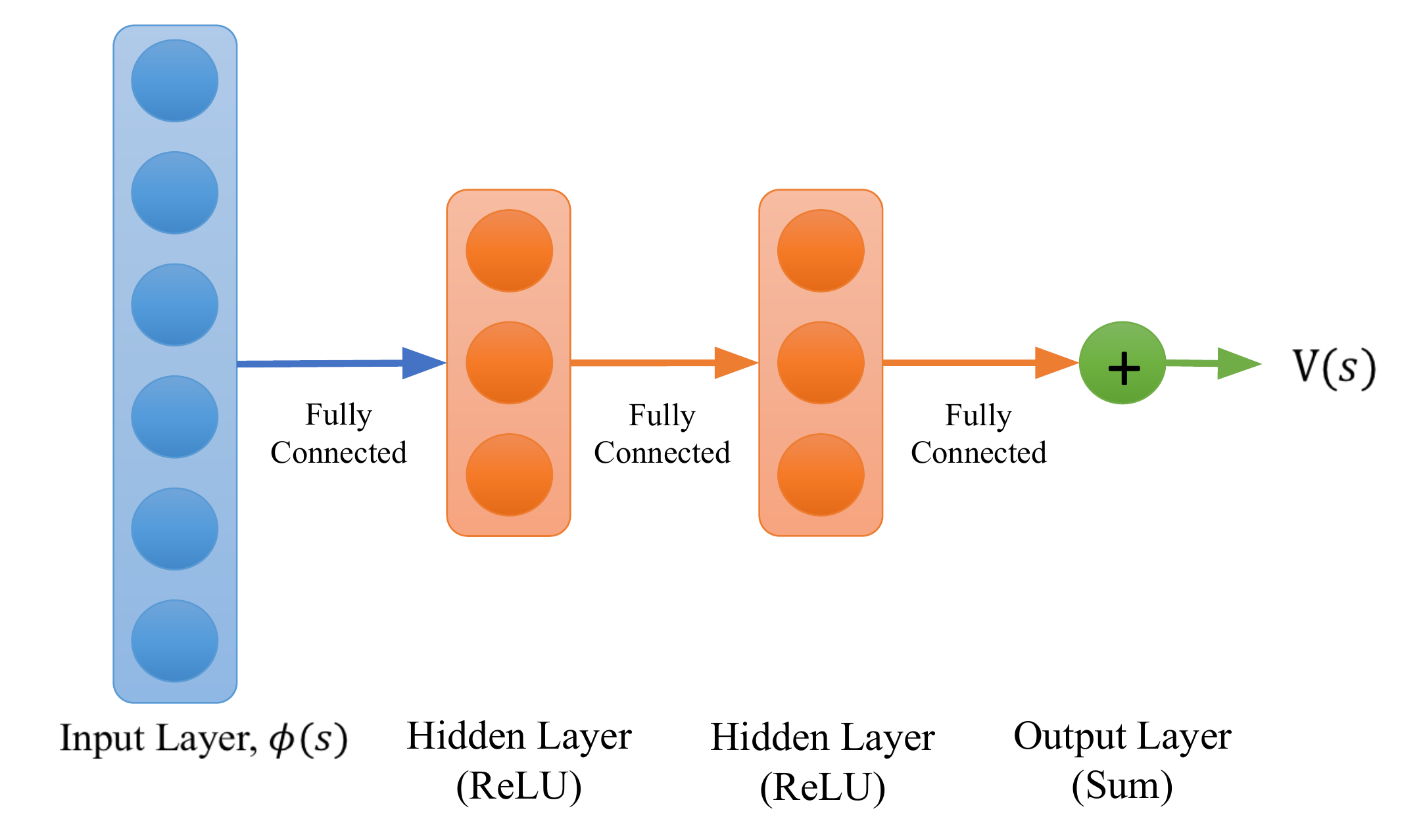}
\caption{The structure of the $V$-values network used in NTD.}
\label{fig:deep_td}
\end{figure}

We use the following settings, which yield good performance
in pilot experiments:
Learning episode budget $T=3000$ and
learning step $\alpha = .001$ in TD and NTD.
\YG{
For NTD, the input of the $V$-value network is 
the same 200-dimensional draft summary representation 
as in \citep{rioux2014emnlp};
after the input layer, we add a fully connected ReLU 
\citep{DBLP:journals/jmlr/GlorotBB11} layer
with dimension 100 as the first hidden layer; an identical fully connected 
100-dimensional ReLU layer is followed as the second hidden layer;
at last, a linear output layer is used to output the $V$-value.
Fig.\ \ref{fig:deep_td} illustrates the structure of the
$V$-values network.}
We use Adam \citep{DBLP:journals/corr/KingmaB14} 
as the gradient optimiser (line 10 in Alg.\ \ref{alg:deep_td}),
with default setup.
For LSTD, we initialise its square matrix as a 
diagonal matrix and let the diagonal elements 
be random numbers between 0 and 1, as suggested by \citet{lagoudakis2003lspi}.

The rewards we use are \ChM[the]{based on} $U^*$ \ChM[(Eq.\ \eqref{eq:u_star})]{defined in Eq.\ \eqref{eq:u_star}}.
Note that this serves as the upper-bound performance, because
$U^*$ is the gold-standard
scoring function, which is not accessible
in the interactive settings (see \S\ref{subsec:full_system_compare}
and \S\ref{sec:final_user_study}).
We measure the performance of the three RL algorithms by 
the quality of their generated summaries
in terms of multiple ROUGE scores.
Results are presented in Table~\ref{table:rl_perf}.
NTD outperforms the other two RL algorithms by a large margin\ChM[,
and we believe]{. We assume} that this is attributed to its more precise
approximation of the $V$-values using the 
neural network.

\begin{table}
  \centering\small
    \begin{tabular}{l l l l l l}
      \toprule
      Dataset & RL &  $R_1$ & $R_2$ & $R_\mathrm{L}$ & $R_\mathrm{SU4}$  \\
      \midrule
      \multirow{3}{*}{DUC'01} & 
      NTD & \textbf{.452}$^*$ & \textbf{.169}$^*$ & 
      \textbf{.359}$^*$  & \textbf{.177} \\
      & TD  & .442 & .161 & .349 & .172 \\
      & LSTD & .432 & .151 & .362 & .179 \\
      \midrule
      \multirow{3}{*}{DUC'02} & 
      NTD & \textbf{.483}$^*$ & \textbf{.181} & 
      \textbf{.379} & \textbf{.193} \\
      & TD & .475 & .179 & .374 & .189 \\
      & LSTD & .462 & .163 & .363 & .183 \\
      \midrule
      \multirow{3}{*}{DUC'04} & 
      NTD & \textbf{.492}$^*$ & \textbf{.189}$^*$ & 
      \textbf{.391}$^*$ & \textbf{.203}$^*$ \\
      & TD  & .473 
      & .174 & .378 & .192 \\
      & LSTD & .457 & .156 & .360 & .182 \\
      \bottomrule
    \end{tabular}
  \caption{NTD outperforms the other TD algorithms
  across all DUC datasets. All results
  are averaged over 10 independent runs across
  all topics in each dataset. Asterisk: significant advantage.}
  \label{table:rl_perf}
\end{table}

In terms of computation time,\footnote{All
RL experiments were performed on a workstation
with a quad-core CPU and 8 GB RAM, without using GPUs.}
TD takes around 30 seconds to finish the
3,000 episodes of training and produce a summary,
NTD takes around 2 minutes, while
LSTD takes around 5 minutes. 
Since the RL computation is performed only once after
all online interaction \ChM[finishes]{has finished}, we find
this computation time acceptable.
However, without using $D_S$ as the memory
replay, NTD takes around 10 minutes to run 3,000 episodes
of training.


\subsection{Full System Performance}
\label{subsec:full_system_compare}
We compare SPPI with two variants of APRIL:
\ChM[APRIL with TD (April-TD)
and with NTD (April-NTD)]{APRIL-TD and APRIL-NTD that use TD and NTD, respectively}.
Both implementations of APRIL use
the \ChM[divergence]{diversity}-based AL strategy (i.e. $div=1.0$).
All the other parameters values are the same
as those described in \S\ref{subsec:query_strategy_compare}
and \S\ref{subsec:rl_comparison}
(see Table~\ref{table:parameters}).



\begin{table}
    \centering
    \small
    \begin{tabular}{ l l l l l l}
    \toprule 
    & &  $R_1$ & $R_2$ & $R_\mathrm{L}$ & $R_\mathrm{SU4}$ \\
    \midrule
    \multirow{3}{*}{$N=0$} & 
    SPPI & .323 & .068 & .259 & .098  \\
    & APRIL-TD & .324 & .070 & .257 & .099\\
    & APRIL-NTD & .325 & .069 & .260 & .100\\
    \midrule
    \multirow{3}{*}{$N=10$} & 
    SPPI & .323 & .068 & .259 & .099 \\
    & APRIL-TD & .338$^*$ & \textbf{.075$^*$} & .268$^*$ & .105$^*$\\
    & APRIL-NTD & \textbf{.339$^*$} & \textbf{.075$^*$} & \textbf{.269$^*$} & \textbf{.106$^*$} \\
    \midrule 
    \multirow{3}{*}{$N=50$} & 
    SPPI & .325 & .067 & .261 & .099 \\
    & APRIL-TD & .340$^*$ & .081$^*$ & .271$^*$ & .106$^*$\\
    & APRIL-NTD & \textbf{.345$^\dagger$} & \textbf{.082$^*$} & \textbf{.276$^\dagger$} & \textbf{.107$^*$}\\
    \midrule
    \multirow{3}{*}{$N=100$} & 
    SPPI & .325 & .070 & .261 & .100 \\
    & APRIL-TD & .349$^*$ & .083$^*$ & .275$^*$ & .113$^*$\\
    & APRIL-NTD & \textbf{.357$^\dagger$} & \textbf{.086$^*$} & \textbf{.281$^\dagger$} & \textbf{.115$^*$} \\
    \bottomrule
    \end{tabular}
    \caption{Results with $N$ rounds
    of interaction with the LNO-based simulated user.
    All results are averaged over all document clusters in DUC'01.
    Asterisk: significantly outperforms SPPI.
    Dagger: significantly outperforms both SPPI and APRIL-TD.}
    \label{tab:results_duc}
\end{table}

Results on DUC'01 are presented in Table~\ref{tab:results_duc}.
When no interaction is allowed (i.e. $N=0$\ChM[), $\Hat{U}=h$
and]{, $\Hat{U}=h$),} we find \ChM{that} the performance of the three algorithms 
\ChM[have]{shows} no significant differences.
With the increase of $N$, the gap between both APRIL
implementations and SPPI becomes larger, 
suggesting the advantage of APRIL over SPPI.
Also note that, when $N=0$ and $N=10$,
\ChM[April]{APRIL}-NTD does not have significant advantage over
\ChM[April]{APRIL}-TD, but when $N \ge 50$, \ChM[April]{APRIL}-NTD significantly
outperforms APRIL-TD 
in terms of \ChM[Rouge]{ROUGE}-1 and \ChM[Rouge]{ROUGE}-L.
This is because when $N$ is small, 
the learnt reward function $\Hat{U}$ contains
much noise (i.e. has low correlation with $U^*$; see 
Table \ref{table:query_perf}) and 
the poor quality of $\Hat{U}$ limits
the advantage of the NTD algorithm. The problem gets relieved 
with the increase of $N$.
The above observations also apply to
the experiments on DUC'02 and DUC'04;
their results are presented in Tables~\ref{tab:appendix:results_duc2}
and Tables~\ref{tab:appendix:results_duc4}, respectively.

\begin{table}[t]
    \centering
    \small
    \begin{tabular}{ l l l l l l}
    \toprule 
    & &  $R_1$ & $R_2$ & $R_\mathrm{L}$ & $R_\mathrm{SU4}$ \\
    \midrule
    \multirow{3}{*}{$N=0$} & 
    SPPI & .350 & .077 & .278 & .112  \\
    & April-TD & .351 & .078 & .278 & .113 \\
    & April-NTD & .350 & .078 & .279 & .112 \\
    \midrule
    \multirow{3}{*}{$N=10$} & 
    SPPI & .349 & .076 & .277 & .111 \\
    & April-TD & .359$^*$ & .084$^*$  & .281  & \textbf{.116$^*$} \\
    & April-NTD & \textbf{.361$^*$}  & \textbf{.085$^*$}  & \textbf{.283$^*$}  & \textbf{.116$^*$}  \\
    \midrule 
    \multirow{3}{*}{$N=50$} & 
    SPPI & .351 & .077 & .279 & .112\\
    & April-TD & .361$^*$ & .083$^*$ & .283 & .117$^*$ \\
    & April-NTD & \textbf{.364$^*$}  & \textbf{.086$^*$}  & \textbf{.287$^\dagger$} & \textbf{.118$^*$} \\
    \midrule
    \multirow{3}{*}{$N=100$} & 
    SPPI & .351 & .078 & .277 & .113 \\
    & April-TD & .368$^*$ & .088$^*$ & .290$^*$ & .123$^*$ \\
    & April-NTD & \textbf{.374$^\dagger$} & \textbf{.089$^*$} & \textbf{.295$^\dagger$} & \textbf{.124$^*$} \\
    \bottomrule
    \end{tabular}
    \caption{Results with $N$ rounds
    of interaction with the LNO-based simulated user in DUC'02.
    All results are averaged over all topics in DUC'02.
    Asterisk: significantly outperforms SPPI.
    Dagger: significantly outperforms both SPPI and April-TD.}
    \label{tab:appendix:results_duc2}
\end{table}

\begin{table}[t]
    \centering
    \small
    \begin{tabular}{ l l l l l l}
    \toprule 
    & &  $R_1$ & $R_2$ & $R_\mathrm{L}$ & $R_\mathrm{SU4}$ \\
    \midrule
    \multirow{3}{*}{$N=0$} & 
    SPPI & .372 & .093 & .293 & .125  \\
    & April-TD & .372 & .091 & .293 & .124 \\
    & April-NTD & .373 & .092 & .292 & .125 \\
    \midrule
    \multirow{3}{*}{$N=10$} & 
    SPPI & .373 & .096 & .297 & .126 \\
    & April-TD & .384$^*$ & \textbf{.098}  & .307$^*$ & .133$^*$  \\
    & April-NTD & \textbf{.388$^*$} & \textbf{.098} & \textbf{.310$^*$}  & \textbf{.134$^*$} \\
    \midrule 
    \multirow{3}{*}{$N=50$} & 
    SPPI & .376 & .096 & .300 & .128 \\
    & April-TD & .388$^*$ & .098 & .307$^*$ & .135$^*$ \\
    & April-NTD & \textbf{.396$^\dagger$}  & \textbf{.100$^*$} & \textbf{.313$^\dagger$} & \textbf{.137$^*$}  \\
    \midrule
    \multirow{3}{*}{$N=100$} & 
    SPPI & .381 & .099 & .301 & .132 \\
    & April-TD & .391$^*$ & .101 & .307$^*$ & .137$^*$ \\
    & April-NTD & \textbf{.407$^\dagger$} & \textbf{.104$^*$}  & \textbf{.316$^\dagger$} & \textbf{.141$^*$}  \\
    \bottomrule
    \end{tabular}
    \caption{Results with $N$ rounds
    of interaction with the LNO-based simulated user in DUC'04.
    All results are averaged over all topics in DUC'04.
    Asterisk: significantly outperforms SPPI.
    Dagger: significantly outperforms both SPPI and April-TD.}
    \label{tab:appendix:results_duc4}
\end{table}

\begin{figure}
    \centering
    \includegraphics[width=0.7\textwidth]{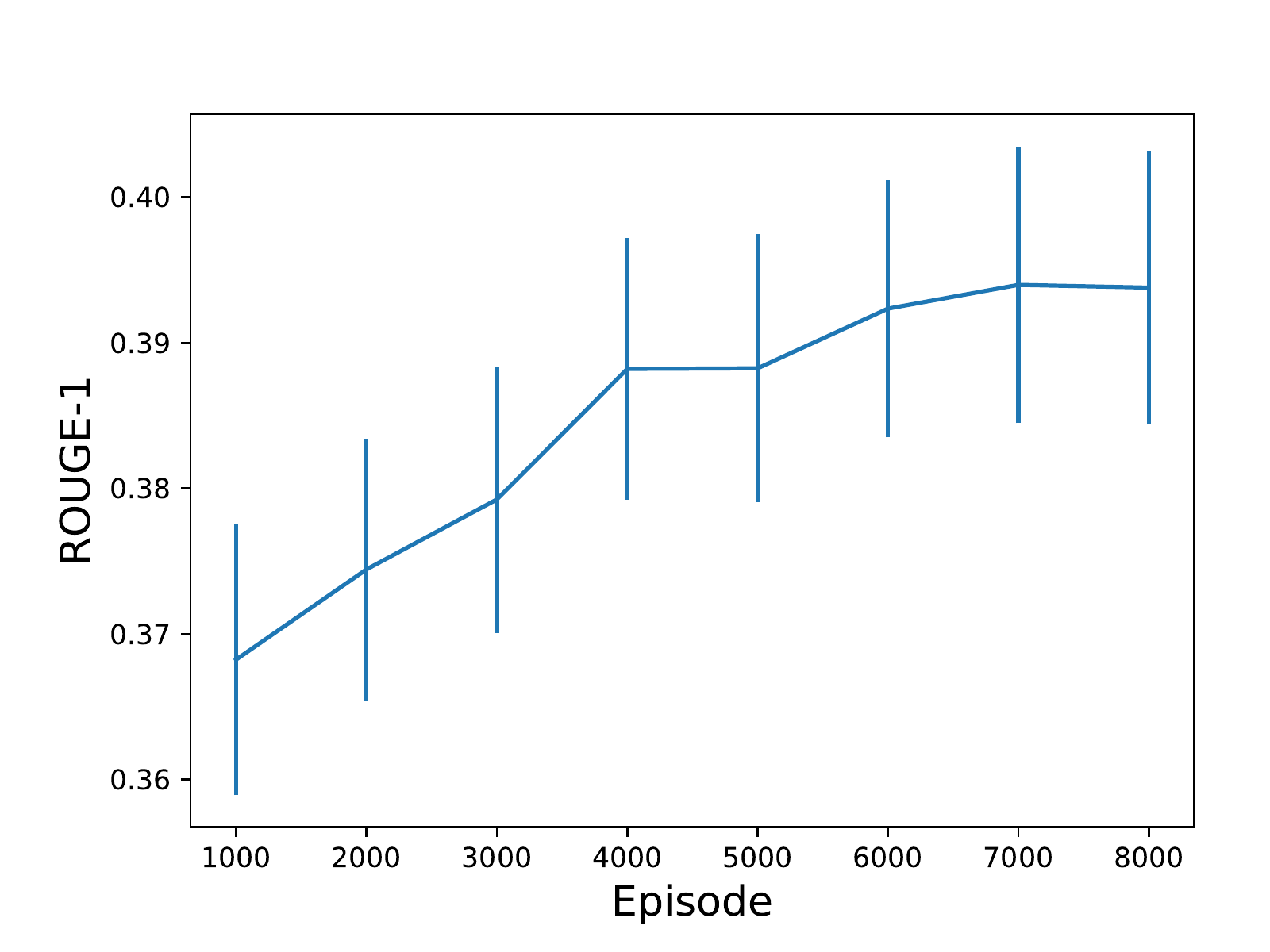}
    \caption{With the same rewards ($\hat{U}$ with $N=300$) 
    and the same RL algorithm (TD), the quality of the 
    generated summary (in terms of ROUGE-1) increases with the
    growth of the episode budget. All results are averaged over 
    all document clusters in DUC'01. Error bars indicate standard errors.}
    \label{fig:episode_rouge}
\end{figure}

We \ChM[believe]{attribute} the superior performance of APRIL \ChM[is
attributed to]{to} two factors:
{(i)} \emph{noise robustness}:
SPPI purely relies
on the collected preferences to improve its policy (see
Alg.\ \ref{alg:sr}),
while our reward estimation considers both collected
preferences as well as heuristics 
to mitigate the noise in the preferences (see Eq.\ \eqref{eq:hat_u}).
{(ii)} \emph{more rounds of training}: 
our method can update its RL policy
for as many episodes as we want ($T \gg N$)
while SPPI can only update its policy for up to $N$ rounds
(see Alg.\ \ref{alg:sr}).
This property enables APRIL to
thoroughly exploit the information from the user preferences.
\YG{
Fig.~\ref{fig:episode_rouge} illustrates that, with the same reward,
the quality of the APRIL-generated summary grows with the increase
of $T$. This is because with more episodes of learning,
the smaller the error between the RL-generated policy and
the optimal policy \citep{Bertsekas96}.
But the computational time is also increased with the 
the growth of $T$: every 1000 episodes costs around 10 seconds
training time for TD, and around 3 minutes for NTD. 
We hence let $T=3000$ (see Table~\ref{table:parameters}) 
in our experiments to trade off between
the performance and real-time responsiveness.
}

\section{Human Evaluation}
\label{sec:final_user_study}

Finally, we invite real users to evaluate the 
performance of APRIL\ChM[. Our experiments had two phases]{\ in two experiments}:
\ChM[\textbf{Phase1} tested]{First, we test} whether APRIL can
improve the no-interaction baseline after
a few rounds of interaction\ChM{\ with the user}.
\ChM[\textbf{Phase2} tested]{Second, we test} whether APRIL outperforms
SPPI \ChM[with]{given} the same query budget.
\ChM[W]{To conduct the experiments, w}e develop a web interface that accommodates
both SPPI and APRIL.
Three document clusters are randomly selected from 
the DUC datasets\ChM{\ (d30046t, d100e and d068f)}.
Seven users participate in our experiments\ChM[ in both phases,
all]{, all of them are native or} fluent in English from our university.
Due to the similar performance of \ChM[April]{APRIL}-TD and \ChM[April]{APRIL}-NTD with $N = 10$ interaction rounds (see Tables~\ref{tab:results_duc},
\ref{tab:appendix:results_duc2} and \ref{tab:appendix:results_duc4}), we use \ChM[April]{APRIL}-TD throughout our experiments, because of its faster computation time (see \S\ref{subsec:rl_comparison}).

\subsection{\ChM[Phase1: APRIL]{APRIL} vs.\ No-Interaction}
\label{subsec:humeval:interaction}

\ChM{Following the setup of our user study introduced in \S\ref{sec:pref_summary}, we first allow the users to understand the background of the topic with two 200-word abstracts.}
Then, we ask them to interact
with APRIL for \ChM[10]{$N = 10$} rounds\ChM[. 
After all interaction finished, we presented]{\ and finally present} 
both the no-interaction summary
(i.e.\ $\beta=0$ in Eq. \eqref{eq:hat_u}) and the
with-interaction summary using \ChM[April]{APRIL}-TD ($\beta=0.5$) 
to the users to ask for their final preference.
Note that, because our AL strategy selects the summaries
to present based on each user's previous preferences,
the summaries read by each user during the interaction
are different.

Table \ref{table:pbrl_human} presents the results.
\ChM{The column ``Human'' shows how often the participants prefer the with-interaction summary over the no-interaction summary.}
For all document clusters we have tested, the users \ChM[prefered]{clearly prefer}  
the with-interaction summaries\ChM[over the no-interaction 
summaries with higher chance, suggesting]{,
suggesting} 
that
APRIL can produce better summaries with 
just 10 rounds of interaction.
In addition, we find that with increasing 
\ChM[$\Delta_{U^*}$ gap]{utility gap $\Delta_{U^*}$} between the
with- and no-interaction summaries, 
the with-interaction summaries
are preferred by more \ChM[(both real and simulated) users.]{users. The column 
``LNO'' compares this finding with our LNO-based user simulation, 
whose probability also increases with $\Delta_{U^*}$.}
This observation \ChM[validates the]{yields further evidence towards our} wider-gap-less-noise assumption 
\ChM[[in \S\ref{sec:pref_summary_study}]{discussed in \S\ref{sec:pref_summary}}. 
Also, the Pearson correlation between the real users'
and LNO's preference ratio
(i.e.\ the second and third column in Table~\ref{table:pbrl_human}) 
is .953 with $p$-value .197, which confirms 
the validity of the LNO-based model. 
However, we also note that in topic d068f, 
the with-interaction summaries'
average $U^*$ is lower than the no-interaction ones,
but \ChM[above 50\% real users still]{still more than 50\% of the users} prefer the 
with-interaction summaries.
We believe \ChM{that} this is due to the mismatch
between the ROUGE-based $U^*$ and users'
\ChM[real measurement over]{real judgement of the} summaries.

\begin{table}
\small
\centering
\begin{tabular}{l c c c}
\toprule
ClusterID & Human & LNO & $\Delta_{U^*_x}$ \\
\midrule
d30046t & 85.7\% & 65.9\% & 1.65\\
d100e & 71.4\% & 60.7\% & 1.08 \\
d068f & 57.1\% & 42.6\% & $-$.75 \\
\bottomrule
\end{tabular}
\caption{Most users prefer the with-interaction
summaries over the no-interaction summaries. Human:
the percentage of pairs in which 
users prefer the with-interaction
over the no-interaction summaries.
LNO: the percentage of pairs in which the 
LNO simulated user prefers
with-interaction. 
$\Delta_{U^*_x}$: the average improvement of the with-interaction
summaries over the no-interaction summaries in terms
of $U^*_x$.
\label{table:pbrl_human}
}
\end{table}

\subsection{\ChM[Phase2: APRIL]{APRIL} vs.\ SPPI}
\label{subsec:humeval:april-sppi}

\begin{figure*}[ht]
\newcommand{\Summary}[2]{\framebox{%
  \begin{minipage}[t][4.4cm][t]{.47\linewidth}%
    \textbf{#1}: {#2}%
  \end{minipage}%
}}
\newcommand{\SummaryShort}[2]{\framebox{%
  \begin{minipage}[t][4.0cm][t]{.47\linewidth}%
    \textbf{#1}: {#2}%
  \end{minipage}%
}}

\scriptsize
\centering
\SummaryShort{Cluster d30046t, SPPI}{
 The famed Allied checkpoint by the Berlin Wall was closed with an elaborate ceremony that brought together the top diplomats from the Germanys and the four World War II Allies. Maik Polster was a stern-faced member of the East German secret police. Checkpoint Charlie, the famed Allied border crossing by the Berlin Wall, was to be hauled away Friday. ``And now, 29 years after it was built, we meet here today to dismantle it and to bury the conflicts it created.'' It was part of my home.'' ``I ran as fast as I could,'' he said. U.S. Army Sgt.
}
\SummaryShort{Cluster d30046t, APRIL}{
 The famed Allied checkpoint by the Berlin Wall was closed with an elaborate ceremony that brought together the top diplomats from the Germanys and the four World War II Allies. ``This is a nice way to end my military service to be here when they take it down,'' said Walsh, 23, a military police officer who leaves the army in six weeks to study for the priesthood. Checkpoint Charlie, the famed Allied border crossing by the Berlin Wall, was to be hauled away Friday. East Germany's border guards were as feared as members of the secret police. U.S. Army Sgt.
}
\Summary{Cluster d100e, SPPI}{
The smart money argues that the Senate could not muster the 67 votes that would be needed to remove the wounded president from office, which would require the defection of 12 Democrats if all the Republicans stand against him. In an incredibly unseemly display, Trent Lott, the majority leader, and former Bush national security adviser Brent Scowcroft and Bush Secretary of State Lawrence Eagleburger chimed in on the attack. Rep. Thomas Barrett, a Democrat from Wisconsin, tried to remind his Republican colleagues that the Constitution ``does not allow you to remove a president from office because you can't stand him.''
}
\Summary{Cluster d100e, APRIL}{
Bob Livingston, the incoming speaker of the House, took no public role Friday as the debate unfolded on whether to impeach President Clinton. ``We 're losing track of distinction between sins and crimes,'' said Rep. Jerrold Nadler, D-N.Y. ``We 're lowering the standards of impeachment. But at the White House, where calls for Clinton's resignation are derided as a Republican strategy, the president sent a spokesman into the driveway to urge Livingston to reconsider his resignation. It has gotten to the point where drastic action may be necessary. The only thing certain now is uncertainty. You resign !''
}
\Summary{Cluster d068f, SPPI}{
Say what you want about Albert Goldman, the author of the new biography, ``The Lives of John Lennon'' (Morrow, \$22.95), but you 've got to hand it to him : This guy is one ambitious sleazemonger. John Lennon's worldwide message of peace was delivered Tuesday as his song ``Imagine'' was played simultaneously for 1 billion people in 130 countries to celebrate what would have been his 50th birthday. The image of a dour, shoeless English boy and his absent, carefree mother prompted Julia Baird and Geoffrey Giuliano to collaborate on a book. ``I believe in fairies, the myths, dragons. Surprised?
}
\Summary{Cluster d068f, APRIL}{
John Lennon's worldwide message of peace was delivered Tuesday as his song ``Imagine'' was played simultaneously for 1 billion people in 130 countries to celebrate what would have been his 50th birthday. The image of a dour, shoeless English boy and his absent, carefree mother prompted Julia Baird and Geoffrey Giuliano to collaborate on a book. Cynthia Lennon joins the throng denouncing the new, unauthorized biography of her late former husband, John Lennon, as written by a money-hungry author capitalizing on untruths. ``I believe in fairies, the myths, dragons. Lennon's widow, Yoko Ono, asked the crowd. Happy birthday, John.
}
\caption{Summaries generated by SPPI and APRIL after 10 rounds of interaction
with real users.}
\label{fig:sppi_april_pairs}
\end{figure*}

\ChM{We invite seven users to judge the quality of SPPI and APRIL summaries in the following way:
We use six randomly selected APRIL-generated with-interaction summaries (two per 
document cluster) from the first experiment (\S\ref{subsec:humeval:interaction}) and pair them with six new SPPI-generated summaries on the same clusters. To generate the SPPI summaries, we ask two additional users to interact with SPPI for ten interaction rounds on the same three document clusters and in the same manner as in the first experiment.
Then, we ask the seven users of the actual study to provide a preference judgement towards the best summary of each pair and additionally rate the quality of each summary on a 5-point Likert scale (higher score means higher quality).}
\YG{Some summaries presented to the users in this user study are
presented in Fig.~\ref{fig:sppi_april_pairs}.}
\ChM{Note that in all previous work we are aware of \citep{avinesh_meyer2017_interactive_doc_sum,DBLP:conf/acl/KreutzerSR17,emnlp18}, the evaluation was based on simulations with a perfect user oracle. Therefore, we expect that our results with real user interaction 
better reflect the true results.}

Table \ref{table:pbrl_sppi} presents the results.
\YG{
In two out of three clusters, the APRIL-generated summaries \ChM[and more preferences from the users]{are clearly preferred by the users and }receive higher
ratings.
The exception is cluster d30046t, where users equally prefer
the SPPI- and APRIL-generated summaries and give them similar ratings.
By looking into \ChM[their]{these} summaries (see the top row in Fig.~\ref{fig:sppi_april_pairs}),
we find that both summaries grasp the main idea of the \ChM{document }cluster
(Checkpoint Charlie is removed with a ceremony, attended by diplomats
from Germany and World War II allies), but also include some
less important information (e.g. ``I ran as fast as I can'' in the summary 
by SPPI, and ``This a nice way ...'' in the one by APRIL).
\ChM[Also note that, again in cluster d30046t, the APRIL-generated summaries 
are overwhelmingly preferred over the non-interaction summaries
(see the top row in Table~\ref{table:pbrl_human}).]{The top row of Table~\ref{table:pbrl_human} suggests that users overwhelmingly prefer APRIL-generated summaries over no-interaction summaries for this cluster d30046t.}
\ChM[These observations suggest that for clusters about a single short-term
event (e.g. d30046t), both interactive approaches can easily grasp the
users' needs for summaries, and hence generate similar quality summaries.]{This suggests that both interactive approach generate summaries of similar high quality for this cluster.
As the cluster is about a single short-term event, we speculate that both interactive approaches can easily grasp the users' needs for such events and produce equally good summaries.}
However, for more complex clusters that are about multiple events (e.g. d100e,
which talks about a series events happened on multiple politicians) 
or about events happened across a long time range (e.g. d068f, which talks about
events before and after the death of John Lennon), 
APRIL can more precisely grasp the need of the users and hence generate
better summaries than SPPI.
}

\YG{
To summarise, given the same query budget $N$, APRIL generates comparable
or superior quality summaries compared to SPPI,}
while its reading load is \ChM[just]{almost} half of SPPI
(\YG{APRIL requires the users reading $N+1$ summaries, while SPPI requires $2N$;} 
see \S\ref{subsec:stage1}). 
Also, we find a high correlation between the real users' and 
the LNO's preference ratio (Pearson correlation
.974 with $p$-value .145), which confirms again the 
validity of LNO.

\begin{table}
\small
\centering
\begin{tabular}{l c c c c}
\toprule
ClusterID & Human & LNO &  $\ChM[S]{Q}_\mathrm{APRIL}$ & $\ChM[S]{Q}_\mathrm{SPPI}$ \\
\midrule
d30046t & 50\% & 40.2\%  & 3.4\phantom{${}^*$}  & 3.2  \\
d100e & 82\% & 75.3\%  & 4.0$^*$  & 2.5  \\
d068f & 75\% & 60.4\% & 3.7$^*$ & 2.3 \\
\bottomrule
\end{tabular}
\caption{Most users prefer the summaries generated by APRIL
over those by SPPI.  
Human: the percentage that users prefer \ChM[PbNRL]{APRIL} over SPPI. 
LNO: the percentage that the LNO-based simulated user
prefers \ChM[PbNRL]{APRIL} over SPPI.
$\ChM[S_x]{Q_m}$: the average ratings of the summaries generated
by method $\ChM[x]{m}$. Asterisk: significant advantage of \ChM[PbNRL]{APRIL} over SPPI.
\label{table:pbrl_sppi}
}
\end{table}

\section{Conclusion}
\label{sec:conclusion}

In this work, we propose a preference-based interactive 
document summarisation framework.
which interactively learns 
to generate improved summaries based on user preferences.
We focused on two research questions in this work,
(i) can users easily provide reliable preferences
over summaries, and (ii) how to mitigate
the high sample complexity problem.
For question (i), we showed in a \ChM[series of user studies]{user study}
that users are more likely to 
provide reliable preferences when the quality gap
between the presented summaries is big, and users find
it is easier to provide preferences than other 
forms of feedback (e.g. bigrams). 
For question (ii), we proposed the APRIL framework,
which splits the reward learning and the 
summary searching stage. This split allows
APRIL to more efficiently query
the user and more thoroughly exploit the
collected preferences by using active 
preference learning algorithms,
and more effectively search for the optimal
summary by using reinforcement learning algorithms.
Both our simulation and real-user experiments 
suggested that, with only a few (e.g. \ChM[10]{ten}) rounds
of interaction, APRIL can generate 
summaries better than \ChM[either the]{the} non-interactive RL-based
summariser \ChM[or]{and} the SPPI-based interactive summariser. 
APRIL has the potential to be applied to
a wide range of other NLP tasks such as machine translation and semantic parsing.



\bibliographystyle{spbasic}      
\bibliography{general_long}

\end{document}